%% file: main.tex
\documentclass[10pt,letterpaper]{article}

\usepackage{cvpr}              

\input{preamble}

\definecolor{cvprblue}{rgb}{0.21,0.49,0.74}
\usepackage[pagebackref,breaklinks,colorlinks,citecolor=cvprblue]{hyperref}

\usepackage{xcolor}
\usepackage[colaction]{multicol}

\def\paperID{6} 
\def\confName{CVPR}
\def\confYear{2024}

\title{SAM-PM: Enhancing Video Camouflaged Object Detection using Spatio-Temporal Attention}

\author{Muhammad Nawfal Meeran \qquad Gokul Adethya T$^\dagger$ \qquad Bhanu Pratyush Mantha$^\dagger$\\
National Institute of Technology, Tiruchirappalli\\
}

\begin{document}
\maketitle
\begin{figure}[!ht]
    \centering
    \includegraphics[width=\textwidth]{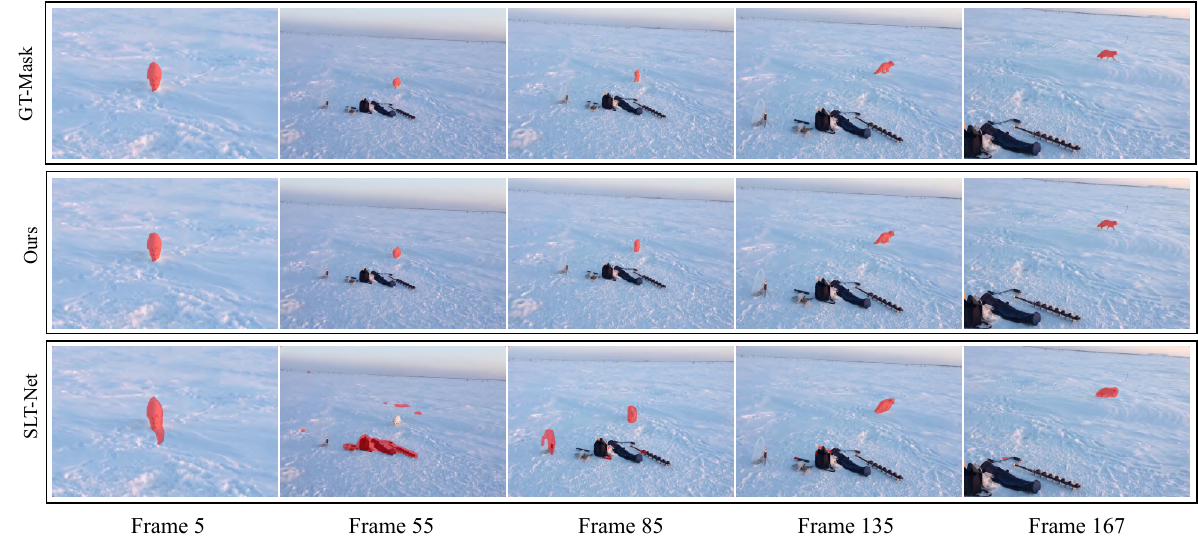}
    \caption{Comparison of mask predictions between ground truth (GT), SAM-PM (Ours), and SLT-Net.} 
    \label{fig:figure1}

\end{figure}.

\begin{multicols}{2}
\input{sec/0_abstract}

\input{sec/1_intro}

\input{sec/2_formatting}
\input{sec/3_finalcopy}
{
    \small
    \bibliographystyle{ieeenat_fullname}
    \bibliography{main}
}
\end{multicols}

\end{document}

%% file: preamble.tex
\usepackage[dvipsnames]{xcolor}
\usepackage{setspace}


%% file: sec/0_abstract.tex
\begin{abstract}
In the domain of large foundation models, the Segment Anything Model (SAM) has gained notable recognition for its exceptional performance in image segmentation. However, tackling the video camouflage object detection (VCOD) task presents a unique challenge. Camouflaged objects typically blend into the background, making them difficult to distinguish in still images. Additionally, ensuring temporal consistency in this context is a challenging problem. As a result, SAM encounters limitations and falls short when applied to the VCOD task. To overcome these challenges, we propose a new method called the SAM Propagation Module (SAM-PM). Our propagation module enforces temporal consistency within SAM by employing spatio-temporal cross-attention mechanisms. Moreover, we exclusively train the propagation module while keeping the SAM network weights frozen, allowing us to integrate task-specific insights with the vast knowledge accumulated by the large model. Our method effectively incorporates temporal consistency and domain-specific expertise into the segmentation network with an addition of less than 1\% of SAM's parameters. Extensive experimentation reveals a substantial performance improvement in the VCOD benchmark when compared to the most recent state-of-the-art techniques. Code and pre-trained weights are open-sourced at \href{https://github.com/SpiderNitt/SAM-PM}{https://github.com/SpiderNitt/SAM-PM}
\end{abstract}

\maketitle
\def\thefootnote{$\dagger$}\footnotetext{
Co-authors who contributed significantly to pivotal ideation.
}\def\thefootnote{\arabic{footnote}}

%% file: sec/1_intro.tex
\section{Introduction}
\label{sec:intro}

The task of detecting objects that seamlessly blend into the background is crucial for various applications, including surveillance and security \cite{liu2019concealed}, autonomous driving \cite{cheng2019noise,ranjan2019competitive}, robotics \cite{michels2005high}, and medical image segmentation \cite{fan2020pranet,wu2021jcs}. This challenging task is addressed by Video Camouflaged Object Detection (VCOD) and Camouflage Object Detection(COD). Despite their broad practical utility, these tasks are daunting as camouflaged objects are often indistinguishable to the naked-eyes. Consequently, in the video domain, related problems like Video Object Segmentation (VOS) \cite{beery2020context,yang2019anchor}, and Video Motion Segmentation (VMS) \cite{ji2014null,yang2021self} have received considerable attention in computer vision, but VCOD remains relatively under-explored.

Recently, the Segment Anything Model (SAM) \cite{kirillov2023segment} has gained a lot of attention among foundation (i.e., models trained on vast amounts of data) models like  CLIP \cite{radford2021learning}, BLIP \cite{li2022blip,li2023blip}, and DALL-E \cite{ramesh2021zero,ramesh2022hierarchical} in the field of computer vision. SAM is capable of producing high-quality segmentation masks in diverse scenarios. However, it performs poorly in VCOD and COD, as indicated by our experiments and the findings of this study \cite{chen2023sam} respectively. In VCOD, SAM confronts three key challenges. Firstly, the extensive visual training corpus primarily comprises objects with well-defined boundaries, neglecting the representation of camouflaged objects characterized by ambiguous and indistinguishable boundaries. Secondly, SAM is trained on static image data and struggles to capture motion and maintain temporal consistency across consecutive video frames. Thirdly, the ambiguous boundary means the appearance of the camouflaged object resembles the background. This gives rise to two fundamental problems: \textbf{1)} the object's boundaries seamlessly merge with the background, becoming perceptible only during movement; \textbf{2)} the object typically exhibits repetitive textures akin to its surroundings. As a result, the determination of pixel movement across frames for motion estimation (e.g., using optical flow) becomes erratic and prone to errors. Consequently, using VOS or VMS methods to address the issue encounters significant failures. In addressing this challenge, the prior approach SLT-Net\cite{cheng2022implicit} has employed two distinct modules to implicitly capture motion and maintain temporal consistency. However, this approach proves to be highly resource-intensive in terms of training, presenting a significant drawback.

In this study, we introduce the \textbf{SAM-Propagation Module} (\textbf{SAM-PM}), a novel approach to tackle VCOD. SAM-PM incorporates a Temporal Fusion Mask Module (TFMM) and a Memory Prior Affinity Module (MPAM) into the Propagation Module (PM) to ensure the temporal consistency of masks through SAM. Utilizing TFMM, we extract mask embeddings that are both spatially and temporally informed. These embeddings play a crucial role in enhancing temporal information within MPAM. Furthermore, to harness the capabilities acquired by SAM from the massive data, we opt to freeze SAM weights and only train our PM to gain the domain-specific information required for VCOD. In this work, our contributions can be summarized as follows:
\begin{itemize}
\item We introduce a novel VCOD framework and demonstrate its effectiveness by deploying it alongside SAM. This approach ensures temporal consistency and imparts domain knowledge through a unified optimization target.
\item Achieving these results is possible by training fewer than 1 million parameters while keeping the SAM weights frozen. This renders our SAM-PM exceptionally parameter-efficient during training.
\item Establishing a new state-of-the-art in the VCOD task, we surpass the performance of the previous state-of-the-art (SOTA) SLT-Net \cite{cheng2022implicit} by an impressive margin.
\end{itemize}

%% file: sec/2_formatting.tex
\begin{figure*}[tb] \centering

    \includegraphics[width=0.875\textwidth,height=0.57\textwidth]{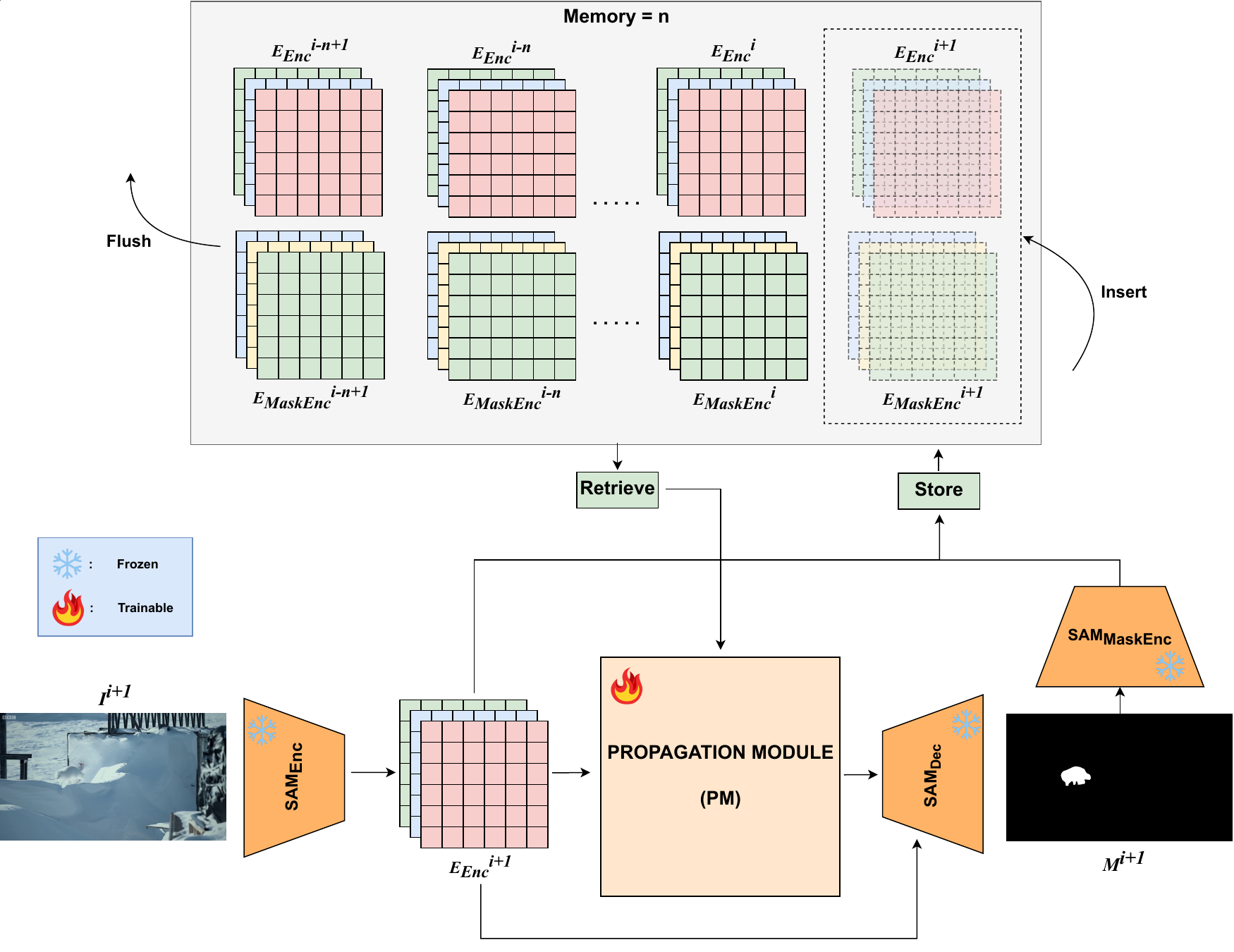}
    \caption{Overall framework of SAM-PM consisting of SAM(frozen), PM(trainable) and Memory. Instead of adding the image and mask directly, we incorporate their embeddings into Memory. This prevents redundant encoding when predicting future masks. One significant advantage of this architecture is that we only need to encode the input frame once, regardless of the number of objects we aim to track.} \label{fig:figure2}
\end{figure*}

\section{Related Works}
\label{sec:formatting}



\textbf{COD.} Even with well-trained eyes, humans can surprisingly overlook camouflaged objects, highlighting the remarkable effectiveness of camouflage in deceiving our visual perception. However, biological studies \cite{hall2013camouflage} reveal that predatory animals follow a two-step process: searching before identifying their prey. Building on this insight, SINet \cite{fan2020camouflaged} introduces two modules: one for the initial search of camouflaged objects and another for their precise detection. Motivated by the same concept, PFNet \cite{mei2021camouflaged} and MirrorNet \cite{yan2021mirrornet} adopt a strategy of first determining the coarse location of camouflage objects and then refining the process through segmentation to generate intricate masks. MGL \cite{zhai2021mutual} incorporated edge details into the segmentation stream via two graph-based modules. More recently, vision transformer-based models like SINet-v2 \cite{fan2022concealed}, ZoomNet \cite{pang2022zoom}, and FSPNet \cite{huang2023feature} have shown strong global and local context modeling capabilities in camouflaged object detection. \\

\noindent\textbf{VCOD.} Prior works \cite{bideau2016s,bideau2018moa,yang2021self} in this field have mainly relied on optical flow-based methods. \cite{meunier2022driven} proposed to segment multiple motions in a rapid and non-iterative way using optical flows. In \cite{yang2021self}, the authors utilized a network to segment the angle field instead of the raw optical flow. \cite{lamdouar2020betrayed} proposed a framework consisting of two components: a registration module aligning background across consecutive frames and a motion segmentation module with memory for detecting moving objects. They also proposed a larger camouflaged dataset (MoCA) with bounding boxes labeled for every five frames. SLT-Net presented a framework that models both short and long-term temporal consistency in camouflaged videos. Furthermore, they proposed a large-scale MoCA-Mask dataset containing 87 video sequences with pixel-wise ground truth masks. 


\noindent\textbf{VOS.}  This task requires localizing objects in videos with temporally consistent pixel-wise masks. Video Object Segmentation (VOS) involves distinct subtasks, including unsupervised (automatic) video segmentation \cite{cho2023treating,lu2019see,wang2017saliency,wang2019learning,wang2019zero}, semi-supervised (mask-guided semi-automatic) video segmentation \cite{lu2020video,yang2022scalable,yang2020collaborative,yang2021associating,yang2021collaborative,yang2022decoupling}, and interactive (scribble or click-based) video segmentation \cite{cheng2021modular}. In the realm of semi-supervised VOS, maintaining temporal consistency has led to the development of several spatiotemporal memory-based methods, such as XMem \cite{cheng2022xmem}, QDMN \cite{liu2022learning}, and Space-Time Memory Network (STM) \cite{oh2019video}. Notably, Space-Time Correspondence Network (STCN) \cite{cheng2021rethinking} introduces a direct image-to-image correspondence approach, which stands out for its simplicity, efficiency, and effectiveness compared to STM. XMem, drawing inspiration from the Atkinson-Shiffrin memory model \cite{atkinson1968human}, proposes a feature memory tailored for long video VOD, addressing the need for temporal coherence. There also has been research done to make VOS more applicable to real-world setting using more complex setups. MOSE \cite{ding2023mose} introduces a diverse dataset featuring numerous crowds, occlusions, and frequent object disappearance-reappearance instances in extended videos. This poses a significant challenge to the performance of existing VOS models. Similarly, other datasets such as OVIS \cite{qi2022occluded} focus on heavily occluded objects, UVO \cite{wang2021unidentified} contributes towards open-world dense objects and VOST \cite{tokmakov2023breaking} is directed toward complex object transformations.

\begin{figure*}[tb] 
    \centering
    \includegraphics[width=0.875\textwidth,height=0.55\textwidth]{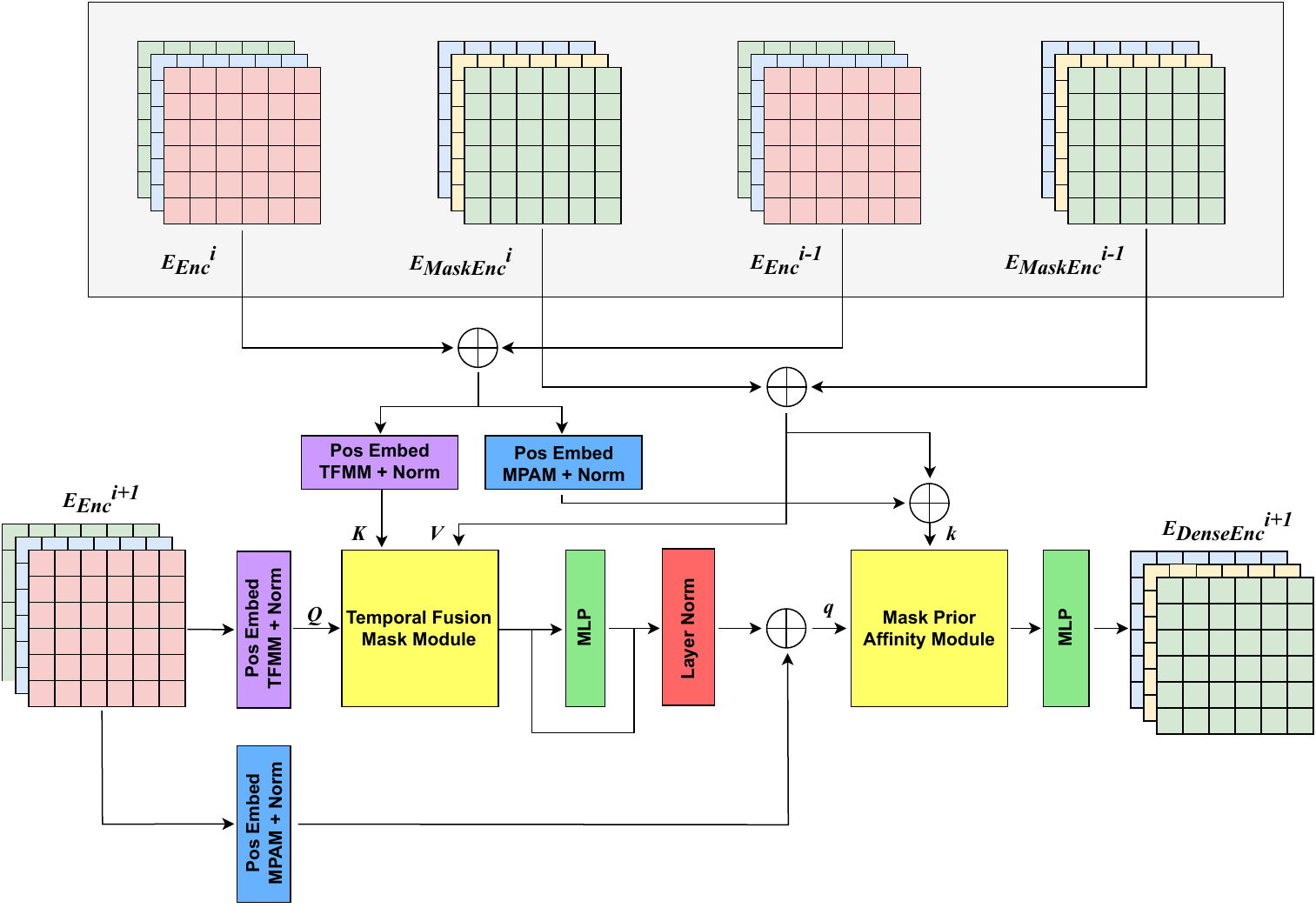}
    \caption{Overview of our Propagation Module consisting of TFMM and MPAM. It's important to observe that we utilize separate positional encoding for each module to provide greater flexibility within the model. Both positional encodings come with trainable parameters, enabling the model to regulate the extent of positional encoding applied to them.} \label{fig:figure3}
\end{figure*}

\section{Proposed Framework}
We propose SAM-PM to adapt SAM for Video Camoulaged Object Detection. In Sec \ref{subsec:Preliminaries_SAM}, we briefly review the architecture of SAM on which our SAM-PM is built. Subsequently, in Sec \ref{subsec:SAM_PM}, we present SAM-PM alongside its key components: the Temporal Fusion Mask Module (TFMM) and the Memory Prior Affinity Module (MPAM), which are the key components to achieve better segmentation quality on videos containing camouflaged objects.

SAM-PM takes as input a video clip featuring camouflaged objects and produces a series of pixel-wise binary masks corresponding to each frame in the video. Let the video clip consisting of $T$ frames be denoted as ${\{I^{i}\}}^{T}_{i=1}$  where $I^{i} \in {R}^{(3\times H\times W)}$. Our objective is to assign binary masks $M^{i} \in {\{0, 1\}}^{(H\times W)}$ to these frames. We encode the frames and generated masks using the SAM image and prompt encoder. We denote the image embeddings and mask embeddings for the $i^{th}$ frame as $E_{Enc}^{i}$ and $E_{MaskEnc}^{i}$ respectively. The initial input to our model is the first frame and the corresponding ground truth.

\subsection{Preliminaries: SAM}\label{subsec:Preliminaries_SAM}
SAM \cite{kirillov2023segment} comprises of three key modules: \textbf{(1)} Image encoder $SAM_{Enc}$: Utilizes a Masked Autoencoder (MAE) \cite{he2022masked} pre-trained Vision Transformer \cite{dosovitskiy2020image} for image feature extraction. The image encoder's output consists of image embeddings 16 times smaller than the original image. \textbf{(2)} Prompt encoder: It encodes the positional information from the input points/boxes to generate sparse embeddings, while input mask prompts are encoded into dense embeddings using the mask encoder $SAM_{MaskEnc}$. \textbf{(3)} Mask decoder $SAM_{Dec}$: It efficiently maps image embeddings and prompt embeddings (i.e. sparse and dense embeddings) to an output mask. The SAM model is trained on the SA-1B dataset, which consists of over 1 billion masks and 11 million images. For more details regarding SAM, we refer readers to \cite{kirillov2023segment}.

\subsection{Ours: SAM-PM}\label{subsec:SAM_PM}
The overall framework of SAM-PM is shown in Figure \ref{fig:figure2}. SAM-PM consists of a Propagation Module (PM) and a simple yet effective Memory alongside the frozen SAM network. SAM-PM addresses the VCOD task in a semi-supervised way, utilizing the first frame's ground truth mask along with video frames as input to prompt encoder and image encoder, respectively. We store the sequence of frames and their corresponding predicted masks in the memory to serve as the inputs to several components within the Propagation Module (PM). The PM is composed of the TFMM and MPAM as shown in Figure \ref{fig:figure3}. The TFMM takes as input the current frame's image embedding, concatenated image embeddings of previous frames, and concatenated predicted mask embeddings from memory. It performs spatio-temporal cross attention, resulting in a temporally infused mask embedding. The MPAM concatenates the temporally infused mask embedding with the current frame's image embedding and the previous frame predicted mask embeddings with their corresponding image embeddings from the memory to perform affinity, consequently enhancing the temporal information. During training, we consider each sample training point to consist of frames extracted from a fixed-length memory $n$, representing subsequences from the full video. During inference, we store and retrieve frames for the whole sequence of the video using a simple strategy. Furthermore, we use a two-stage training method to maintain stability during the training process.

For a given frame $j$ we calculate $E_{Enc}^j$ and $E_{MaskEnc}^j$ as follows
\begin{gather}
E_{Enc}^j= SAM_{Enc}(I^j) \label{eqn1} \\
E_{MaskEnc}^j= SAM_{MaskEnc}(M^j) \label{eqn2}
\end{gather}

\begin{spacing}{1.075}
To predict the mask $M^{i+1}$ for the ${i+1}^{th}$ frame, we first pass the current frame's image embedding $E_{Enc}^{i+1}$, along with the image embeddings $E_{Enc}^{i-n+1:i}$ and the corresponding mask embeddings $E_{MaskEnc}^{i-n+1:i}$ for the previous $n$ frames, to the PM module. This yields the dense embedding $E_{DenseEnc}^{i+1}$ for the current frame as described in Eqn. \ref{eqn4}
\end{spacing}
\begin{gather}
E_{Enc}^{i+1}= SAM_{Enc}(I^{i+1}) \label{eqn3}\\
E_{DenseEnc}^{i+1}= PM(E_{Enc}^{i-n+1:i+1},E_{MaskEnc}^{i-n+1:i}) \label{eqn4}\\
M^{i+1}= SAM_{Dec}(E_{Enc}^{i+1},E_{DenseEnc}^{i+1}) \label{eqn5}
\end{gather}

In Eqn. \ref{eqn5}, the dense and image embeddings of the current frame are passed to the decoder to obtain the current mask prediction. This process is repeated for the entire video, as described in Eqn. \ref{eqn3}-\ref{eqn5} during which the memory is updated by adding new image and dense embeddings while removing outdated ones when the memory limit is reached.

\subsubsection{Temporal Fusion Mask Module}
Building upon prior VOS research \cite{cheng2021rethinking, cheng2022xmem, liu2022learning} emphasizing the incorporation of cross-attention mechanisms for collecting spatio-temporal information within the mask embedding space, we introduce the Temporal Fusion Mask Module (TFMM). The goal of TFMM is to produce mask embeddings that are infused with temporal information from the previous frames.

In Eqn. \ref{eqn6}, we start by adding positional embedding (PE) to the current frame image embedding. Then, we apply LayerNorm (LN) and use the query head of TFMM to obtain $Q$. Similarly, in Eqn. \ref{eqn7}, we add positional embedding to the concatenated image embeddings from memory, apply LayerNorm, and use the key head of TFMM to obtain $K$. In Eqn. \ref{eqn8}, we apply LayerNorm to the concatenated mask embeddings and use the value head of TFMM to obtain $V$.
\begin{gather}
Q= Q_{TFMM}(LN(PE_{TFFM}+E_{Enc}^{i+1})) \label{eqn6}\\
K= K_{TFMM}(LN(PE_{TFFM}+E_{Enc}^{(i-n+1:i)}) \label{eqn7}\\
V= V_{TFMM}(LN(E_{MaskEnc}^{(i-n+1:i)})) \label{eqn8}
\end{gather}

We utilize the standard cross-attention mechanism with $Q$, $K$, and $V$, followed by an MLP (with skip connection) and LayerNorm, to obtain the mask embeddings $Output_{TFMM}$, as described in Eqn. \ref{eqn9}-\ref{eqn11}.
\begin{gather}
O_{TFMM} = Attention_{TFMM}(Q,K,V) \label{eqn9}\\
O_{TFMM} = MLP(O_{TFMM})+ O_{TFMM} \label{eqn10} \\
Output_{TFMM} = LN(O_{TFMM}) \label{eqn11}
\end{gather}

\subsubsection{Memory Prior Affinity Module}
Drawing inspiration from \cite{oh2019video, liu2022learning}, we introduce the MPAM module as an extension. Previous approaches involve using either the sole current image post-image encoder processing \cite{oh2019video} or concatenating the current frame with the prior \cite{liu2022learning} mask before undergoing encoder processing. Instead of relying on preceding masks, we extract values from our Temporal Fusion Mask Module $Output_{TFMM}$. These values offer a generalized representation of mask embeddings that are subsequently concatenated ($\oplus$) with the current frame's image embedding $E_{Enc}^{i+1}$. This enhances information flow compared to utilizing the previous frame in its raw form. This method leverages existing SAM encoders, eliminating the necessity to train an additional encoder as typically needed when concatenating images and masks within the original image space. We use an embedding dimension of 128 for $q_{k}$, $q_{v}$, $m_{k}$ and $m_{v}$ and define them as follows
\begin{gather}
q= LN(PE_{MPAM}+E_{Enc}^{i+1}\oplus Output_{TFMM}))\label{eqn12} \\
k= LN(PE_{MPAM}+E_{Enc}^{(i:i-n+1)})\label{eqn13} \\
q_{k},q_{v} = Linear_{1}(q),Linear_{2}(q)\label{eqn14} \\
m_{k},m_{v} = Linear_{3}(k), Linear_{4}(k)\label{eqn15}
\end{gather}

Using the values defined in Eqn. \ref{eqn14}-\ref{eqn15}, we compute $V_{MPAM}$ and the dense embedding from our module as follows
\begin{gather}
V_{MPAM} = q_{v} \oplus  Attention (q_{k},m_{k},m_{v})\label{eqn16} \\
E_{DenseEnc}^{i+1} = MLP(V_{MPAM})\label{eqn17}
\end{gather}

Finally, we pass both $E_{DenseEnc}^{i+1}$ and $E_{Enc}^{i+1}$ to the decoder $SAM_{Dec}$ to get the predicted mask $M^{i+1}$ as described in Eqn. \ref{eqn5}.


\section{Experiments}
Here, we analyze our proposed framework on the CAD and MoCA-Mask datasets.

\subsection{Datasets}
We use three publicly available camouflage datasets: COD10K \cite{fan2022concealed}, CAD \cite{bideau2016s}, and MoCA-Mask \cite{cheng2022implicit}.

\textbf{COD10K} Currently, it is the largest COD dataset available, which consists of 5066 camouflaged images (3040 for training, 2026 for testing), 1934 non-camouflaged images, and 3000 background images divided into 10 superclasses and 78 sub-classes (69 camouflaged, 9 non-camouflaged).

\textbf{CAD} CamouflagedAnimalDataset consists of 9 video sequences that were extracted from YouTube videos and contain hand-labeled ground-truth masks on every 5th frame.

\textbf{MoCA-Mask} MoCA-Mask dataset consists of 87 video sequences (71 for training, 16 for testing) which extend the bounding box annotations provided by the original Moving Camouflaged Animal (MoCA) dataset \cite{lamdouar2020betrayed} to dense segmentation masks

\end{multicols}
\begin{table*}[tb]
    \caption{Quantitative results on different VCOD Datasets. The best results are highlighted in \textbf{bold}}
    \label{tab:quant_results}
    \centering
    \resizebox{\textwidth}{!}{
    \begin{tabular}{ |l|*{6}{c}|*{6}{c}|*{6}{c}| }
    \hline
     &
    \multicolumn{6}{c|}{MoCA-Mask w/o pseudo labels} & \multicolumn{6}{c|}{MoCA-Mask with pseudo labels} & \multicolumn{6}{c|}{CAD}  \\ \cline{2-19}
    \fontsize{10pt}{12pt}\selectfont Model & \textit{$S_{\alpha}\uparrow$} & \textit{${F_{\beta}^{w}}\uparrow$} & \textit{$E_{\phi}\uparrow$} & \textit{$M\downarrow$} & mDic & mIoU & \textit{$S_{\alpha}\uparrow$} & \textit{${F_{\beta}^{w}}\uparrow$} & \textit{$E_{\phi}\uparrow$} & \textit{$M\downarrow$} & mDic & mIoU & \textit{$S_{\alpha}\uparrow$} & \textit{${F_{\beta}^{w}}\uparrow$} & \textit{$E_{\phi}\uparrow$} & \textit{$M\downarrow$} & mDic & mIoU \\
    \hline
  
    EGNet \cite{zhao2019egnet} & 0.547 & 0.110 & 0.574 & 0.035 & 0.143 & 0.096 & 0.546 & 0.105 & 0.573 & 0.034 & 0.135 & 0.090 & 0.619 & 0.298 & 0.666 & 0.044 & 0.324 & 0.243 \\
    
    BASnet \cite{qin2019basnet} & 0.561 & 0.154 & 0.598 & 0.042 & 0.190 & 0.137 & 0.537 & 0.114 & 0.579 & 0.045 & 0.135 & 0.100 & 0.639 & 0.349 & 0.773 & 0.054 & 0.393 & 0.293 \\
    
    CPD \cite{wu2019cascaded} & 0.561 & 0.121 & 0.613 & 0.041 & 0.162 & 0.113 & 0.550 & 0.117 & 0.613 & 0.038 & 0.147 & 0.104 & 0.622 & 0.289 & 0.667 & 0.049 & 0.330 & 0.239 \\
    
    PraNet \cite{fan2020pranet} & 0.614 & 0.266 & 0.674 & 0.030 & 0.311 & 0.234 & 0.568 & 0.171 & 0.576 & 0.045 & 0.211 & 0.152 & 0.629 & 0.352 & 0.763 & 0.042 & 0.378 & 0.290 \\
    
    SINet \cite{fan2020camouflaged} & 0.598 & 0.231 & 0.699 & 0.028 & 0.276 & 0.202 & 0.574 & 0.185 & 0.655 & 0.030 & 0.221 & 0.156 & 0.636 & 0.346 & 0.775 & 0.041 & 0.381 & 0.283 \\
    
    SINet-v2 \cite{fan2022concealed} & 0.588 & 0.204 & 0.642 & 0.031 & 0.245 & 0.180 & 0.571 & 0.175 & 0.608 & 0.035 & 0.211 & 0.153 & 0.653 & 0.382 & 0.762 & 0.039 & 0.413 & 0.318 \\
    
    \hline
    
    PNS-Net \cite{ji2021progressively} & 0.544 & 0.097 & 0.510 & 0.033 & 0.121 & 0.101 & 0.576 & 0.134 & 0.562 & 0.038 & 0.189 & 0.133 & 0.655 & 0.325 & 0.673 & 0.048 & 0.384 & 0.290 \\
    
    RCRNet \cite{yan2019semi} & 0.555 & 0.138 & 0.527 & 0.033 & 0.171 & 0.116 & 0.597 & 0.174 & 0.583 & 0.025 & 0.194 & 0.137 & 0.627 & 0.287 & 0.666 & 0.048 & 0.309 & 0.229 \\
    
    MG \cite{yang2021self} & 0.530 & 0.168 & 0.561 & 0.067 & 0.181 & 0.127 & 0.547 & 0.165 & 0.537 & 0.095 & 0.197 & 0.141 & 0.594 & 0.336 & 0.691 & 0.059 & 0.368 & 0.268 \\
    
    SLT-Net \cite{cheng2022implicit} & 0.631 & 0.311 & 0.759 & 0.027 & 0.360 & 0.272 & 0.656 & 0.357 & \textbf{0.785} & 0.021 & 0.397 & 0.310 & 0.696 & 0.481 & \textbf{0.845} & 0.030 & 0.493 & 0.401 \\
    
    \hline
    
    SAM \cite{kirillov2023segment} & 0.667 & 0.547 & 0.733 & 0.138 & 0.559 & 0.469 & 0.650 & \textbf{0.517} & 0.722 & 0.140 & \textbf{0.537} & \textbf{0.441} & 0.490 & 0.211 & 0.460 & 0.184 & 0.214 & 0.152 \\
    
    \textbf{SAM-PM} (Ours) & \textbf{0.728} & \textbf{0.567} & \textbf{0.813} & \textbf{0.009} & \textbf{0.594} & \textbf{0.502} & \textbf{0.695} & 0.464 & 0.732 & \textbf{0.011} & 0.497 & 0.416 & \textbf{0.729} & \textbf{0.602} & 0.746 & \textbf{0.018} & \textbf{0.594} & \textbf{0.493} \\
    \hline
    \end{tabular} }
\end{table*}

\begin{figure*}[tb] \centering
    \includegraphics[width=0.120\textwidth]{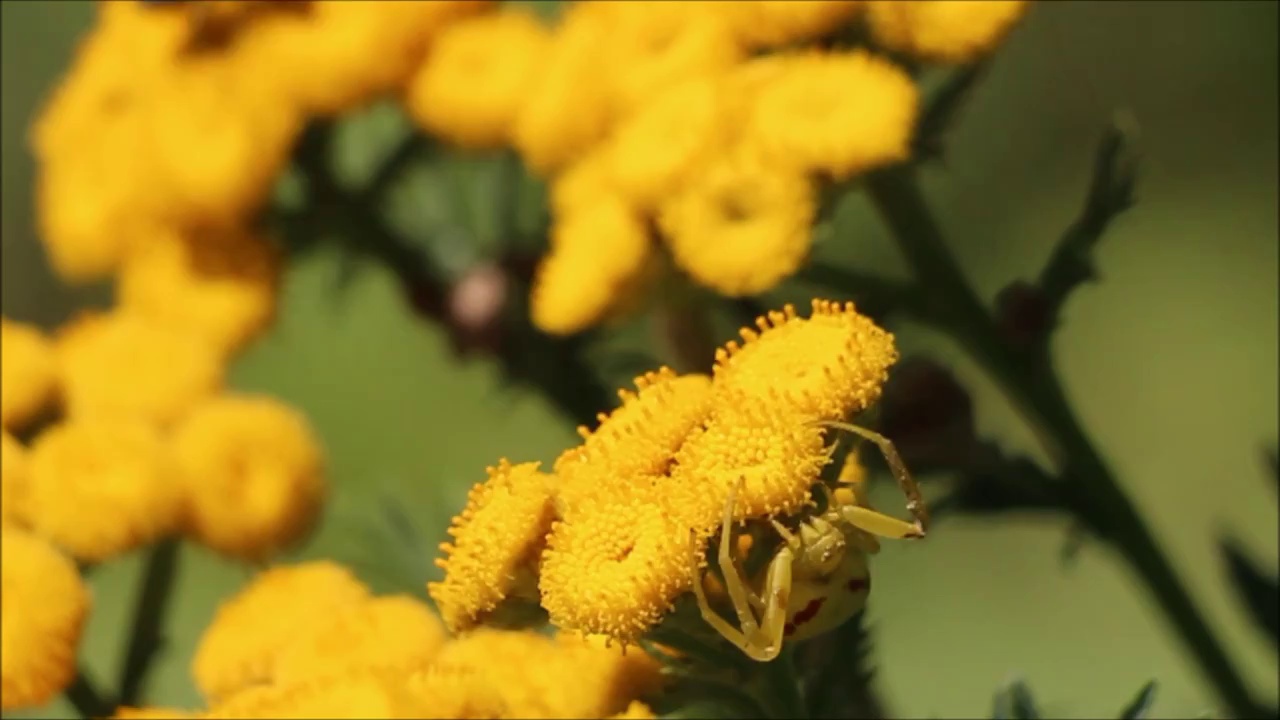}
    \includegraphics[width=0.120\textwidth]{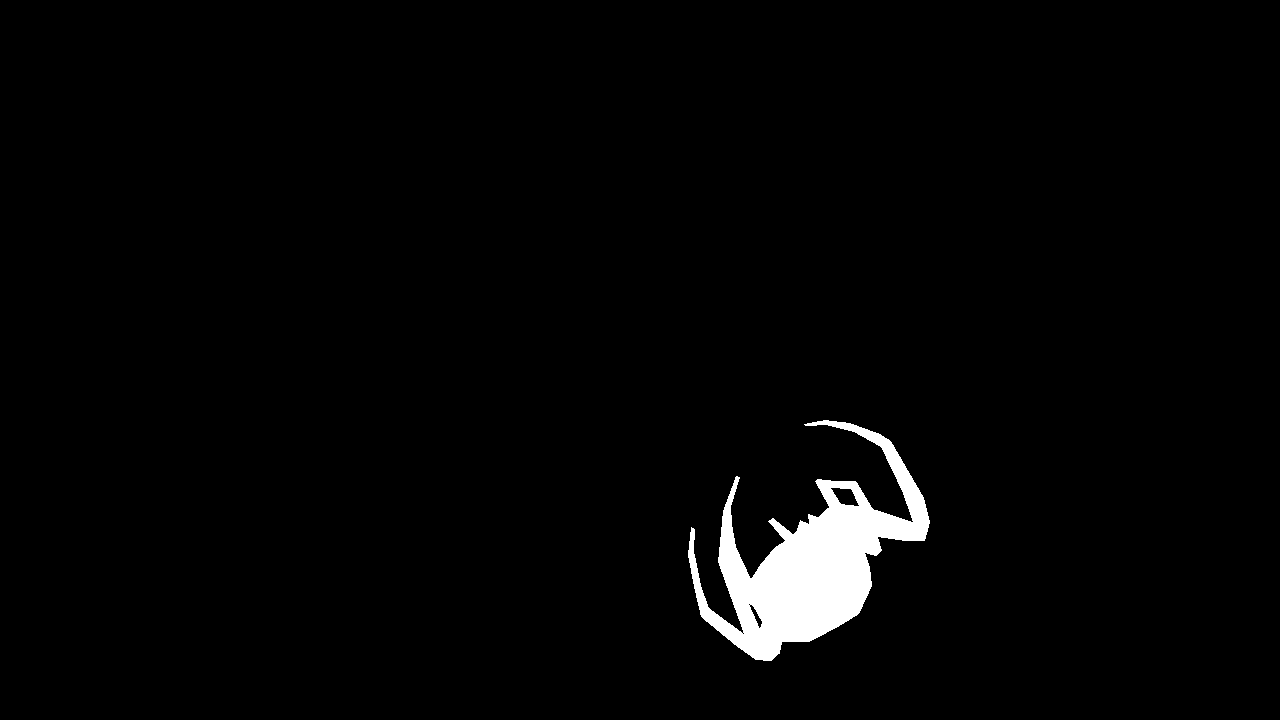}
    \includegraphics[width=0.120\textwidth]{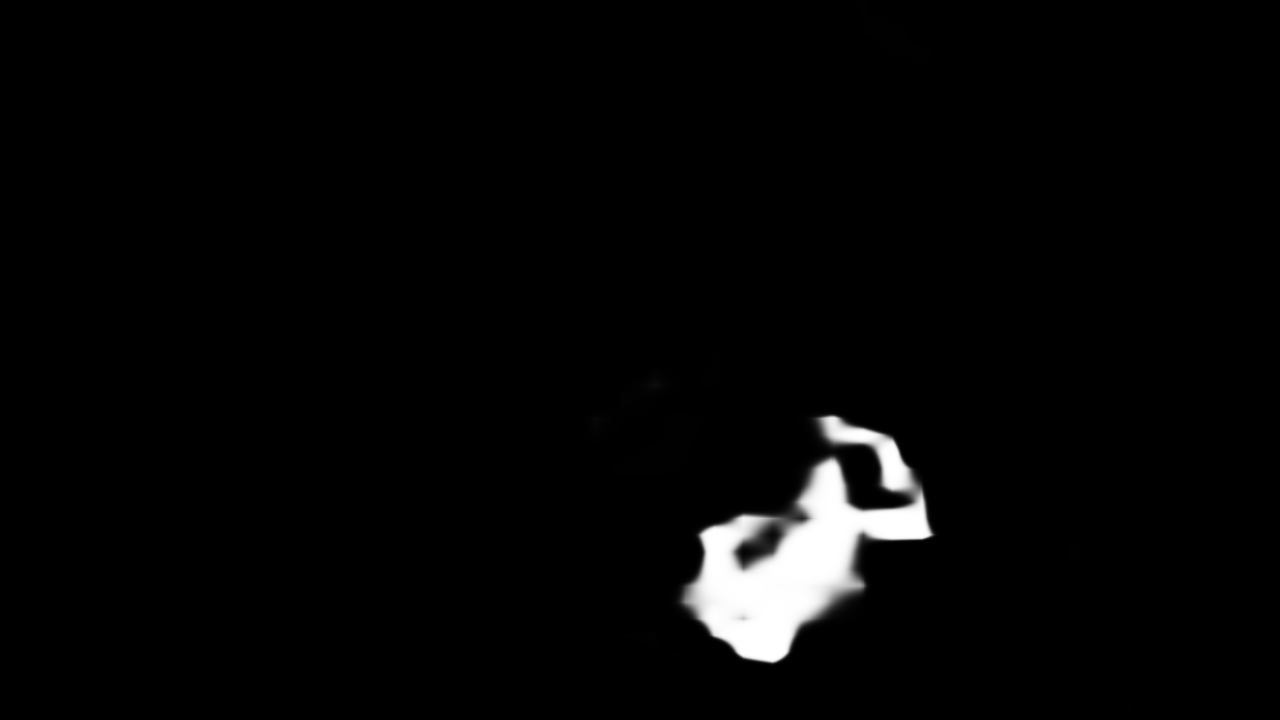}
    \includegraphics[width=0.120\textwidth]{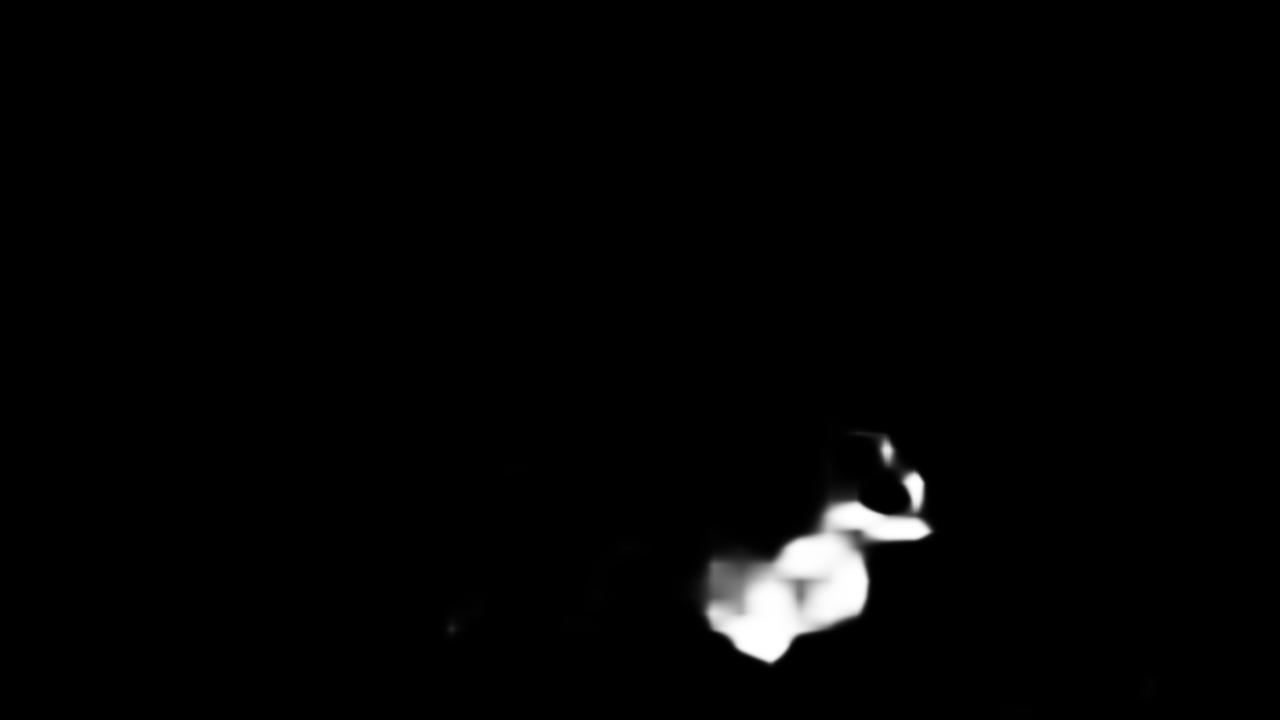}
    \includegraphics[width=0.120\textwidth]{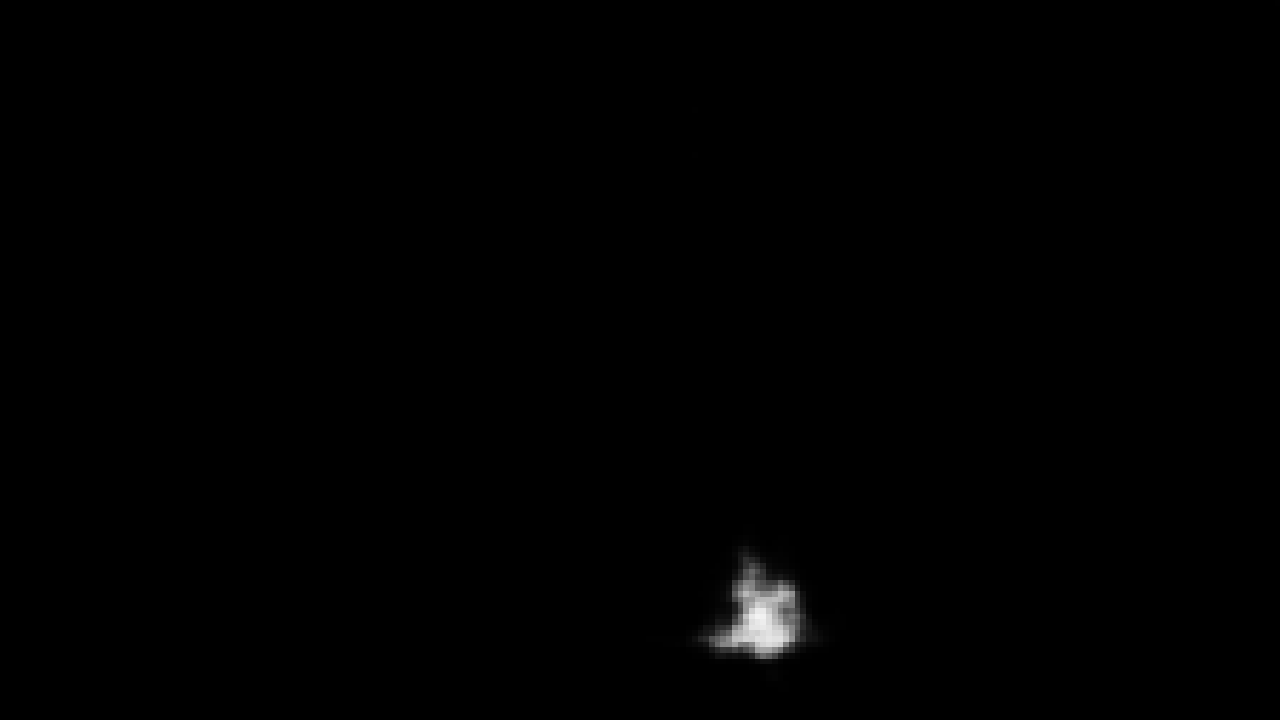}
    \includegraphics[width=0.120\textwidth]{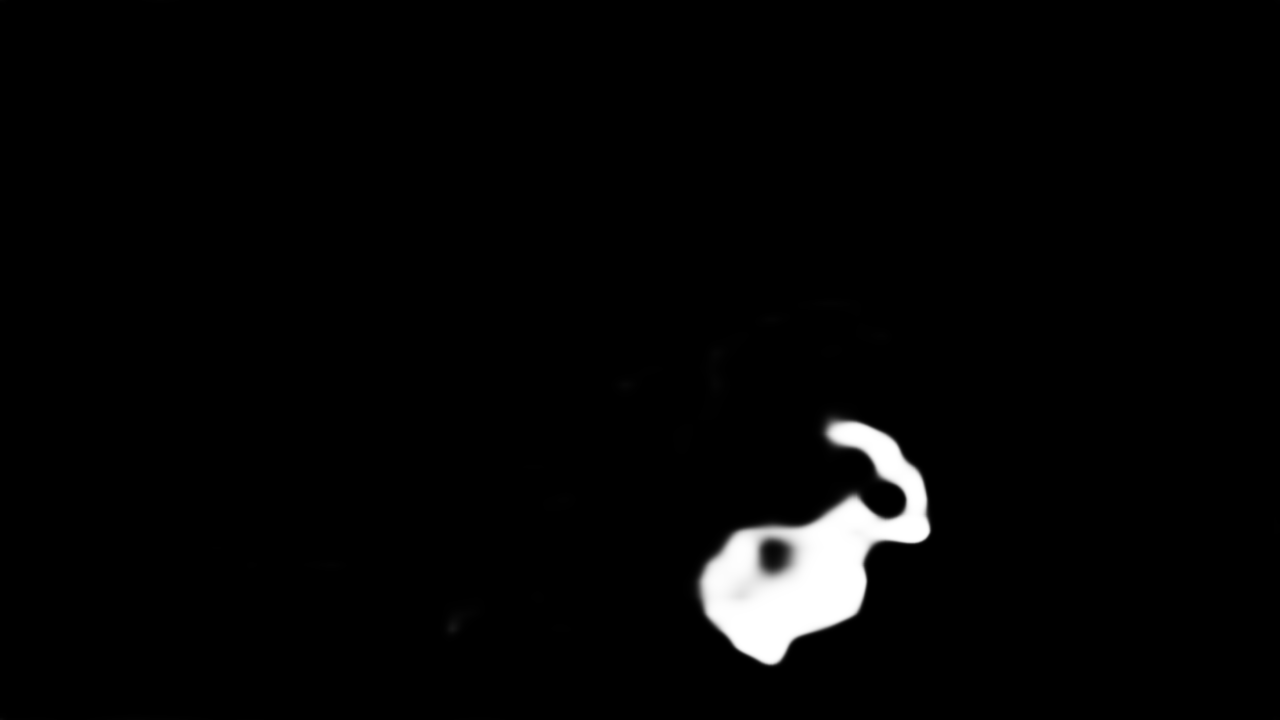}
    \includegraphics[width=0.120\textwidth]{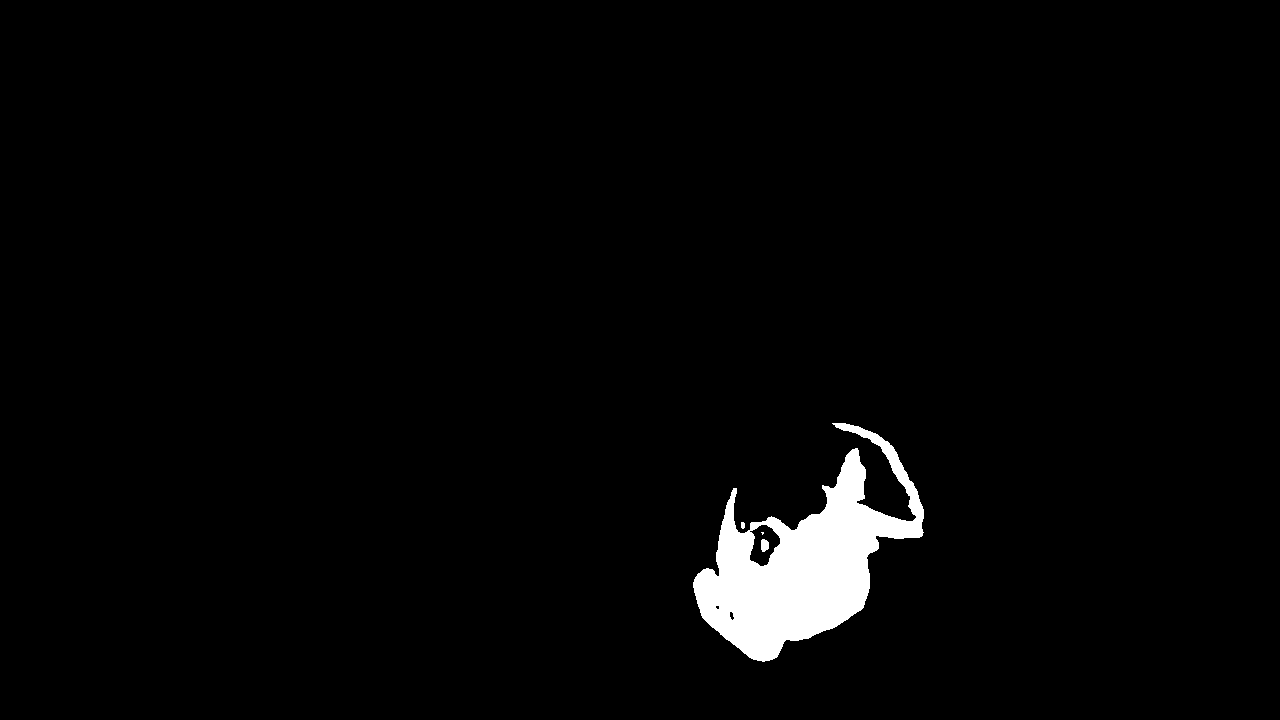}
    \includegraphics[width=0.120\textwidth]{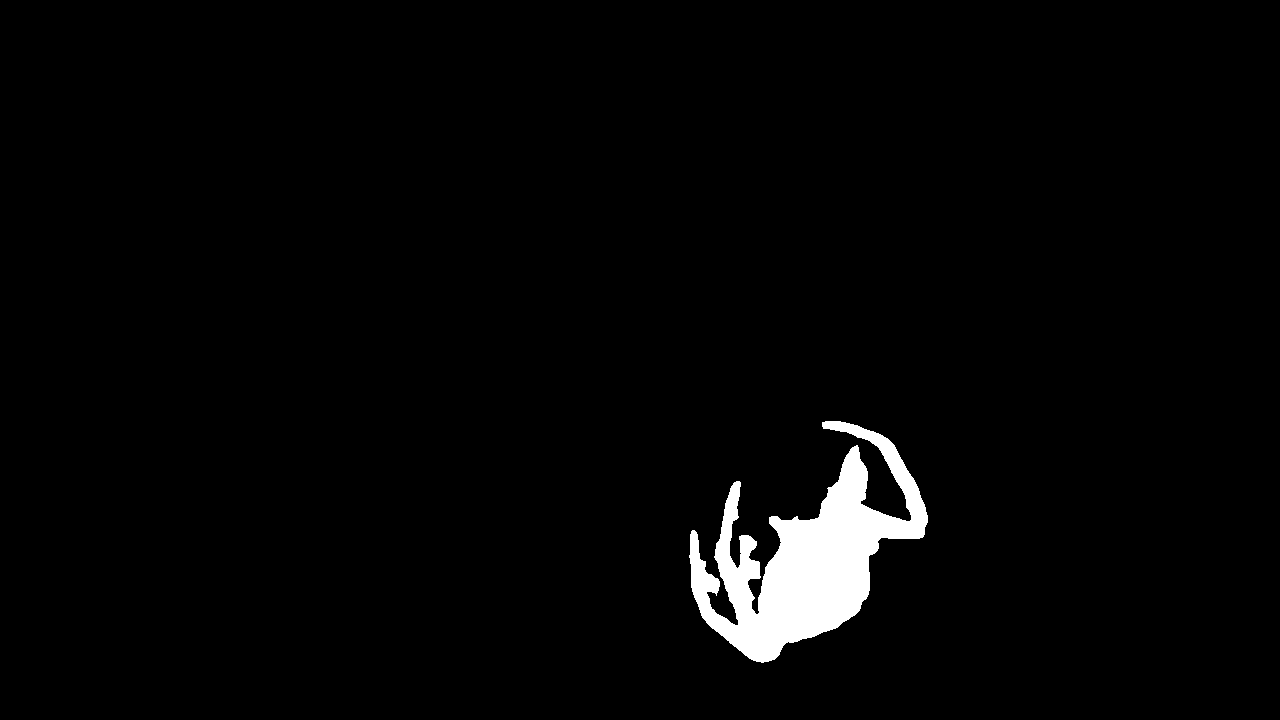}
    \\
    \includegraphics[width=0.120\textwidth]{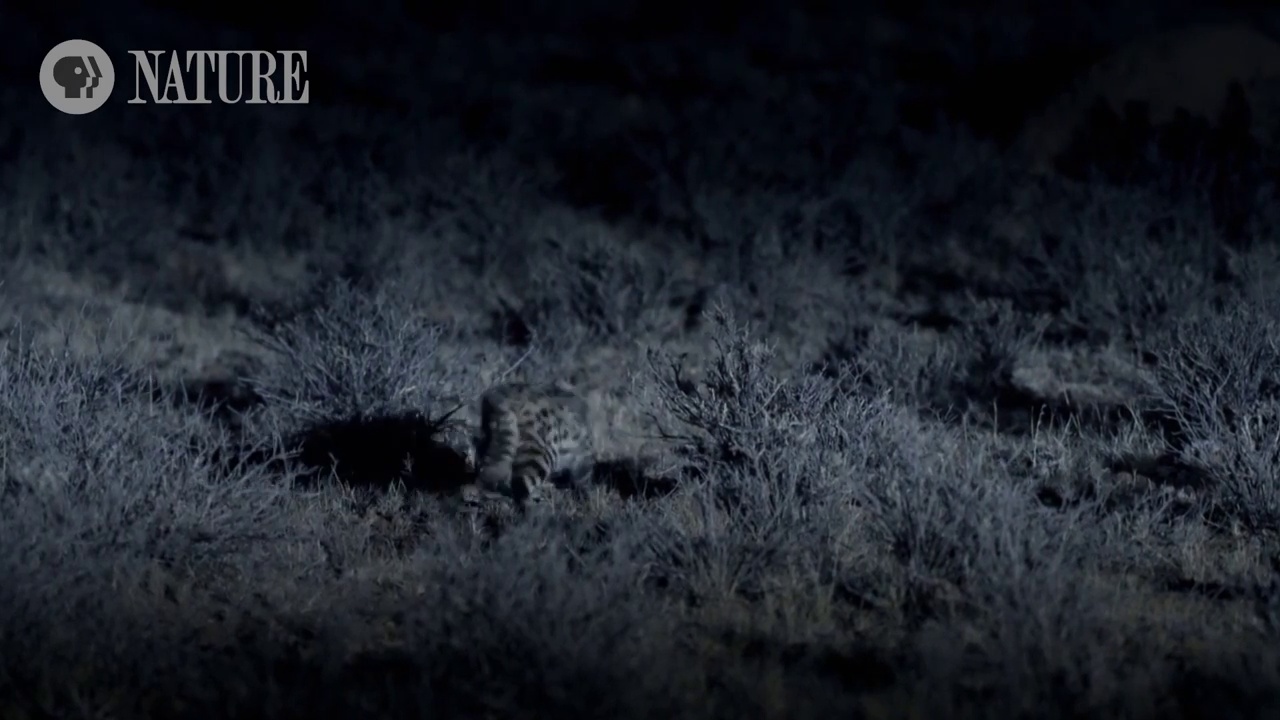}
    \includegraphics[width=0.120\textwidth]{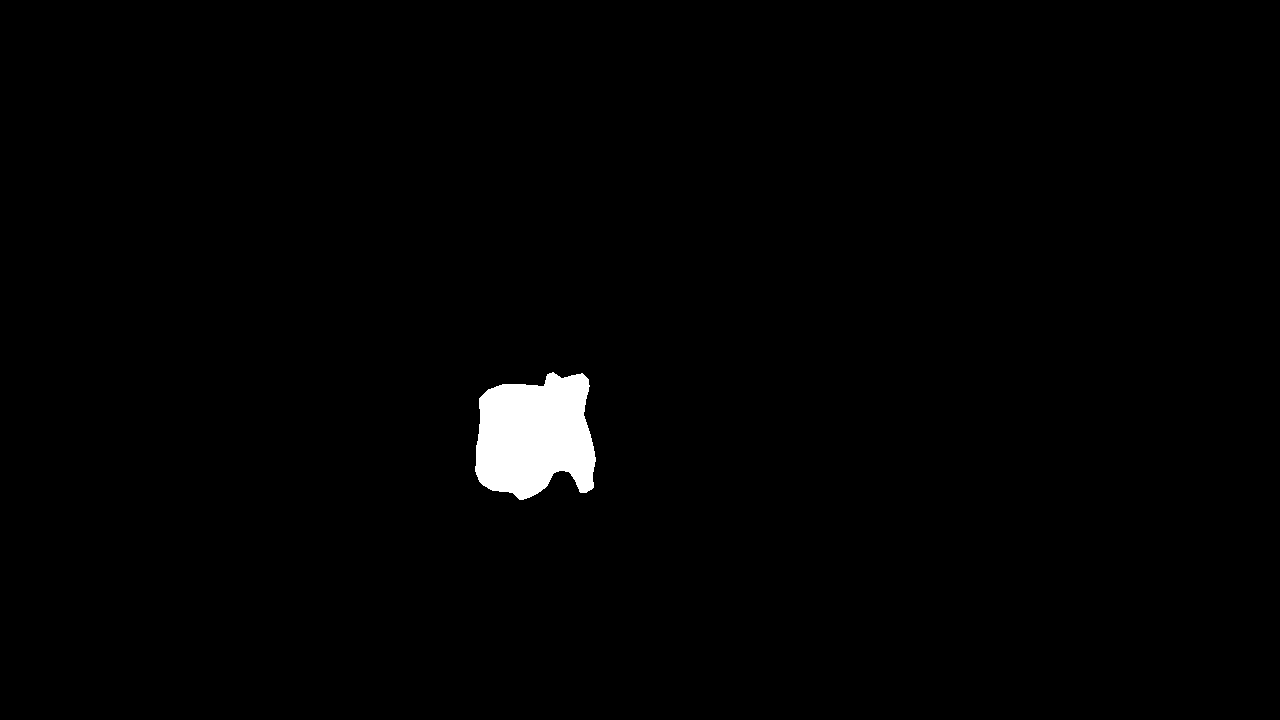}
    \includegraphics[width=0.120\textwidth]{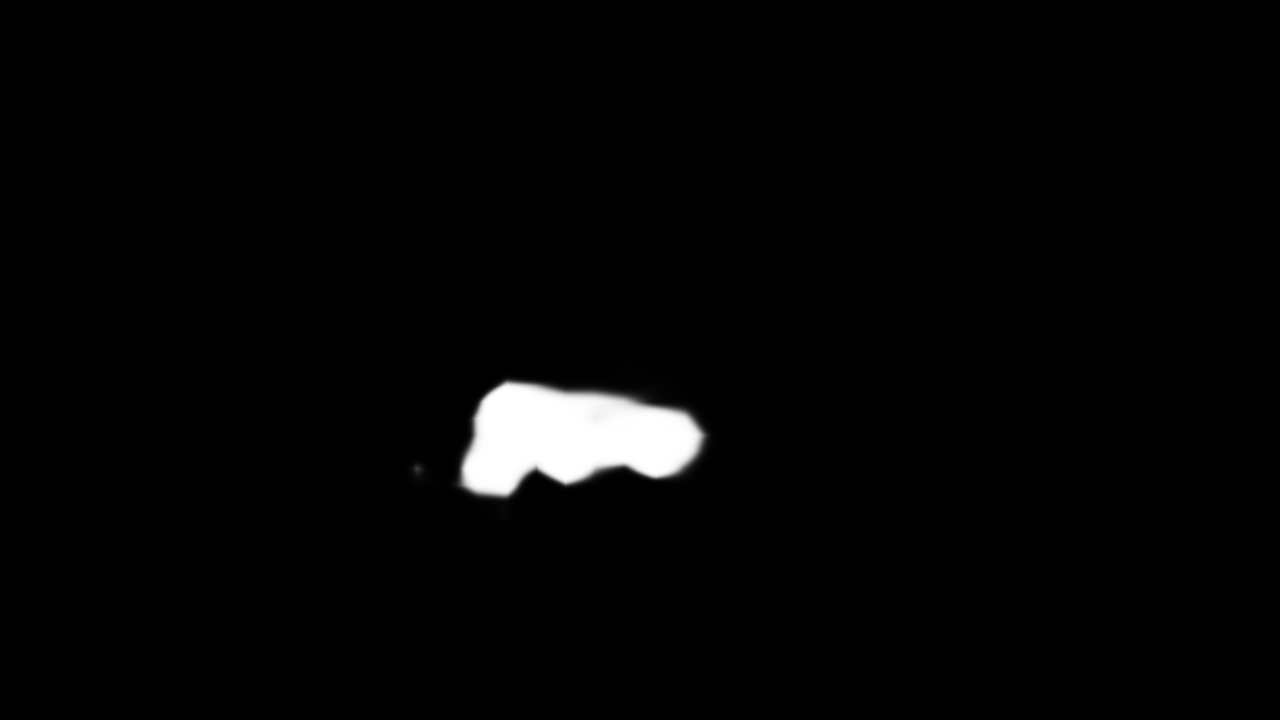}
    \includegraphics[width=0.120\textwidth]{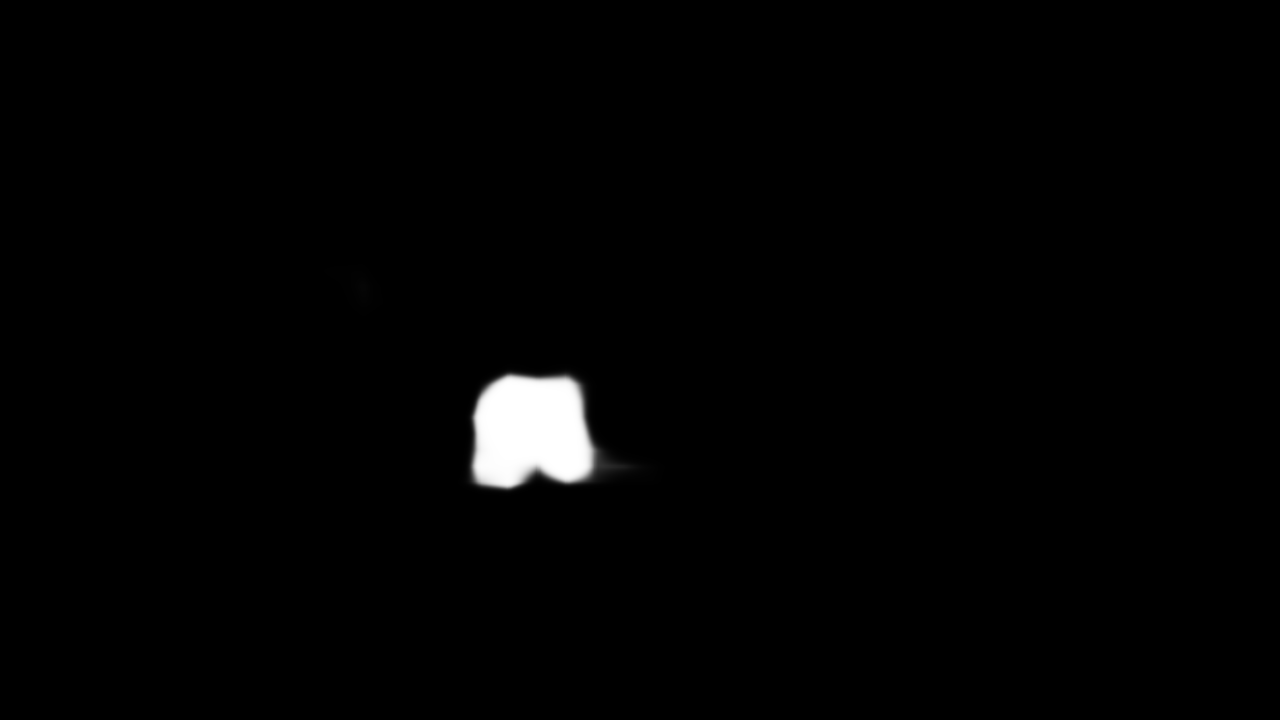}
    \includegraphics[width=0.120\textwidth]{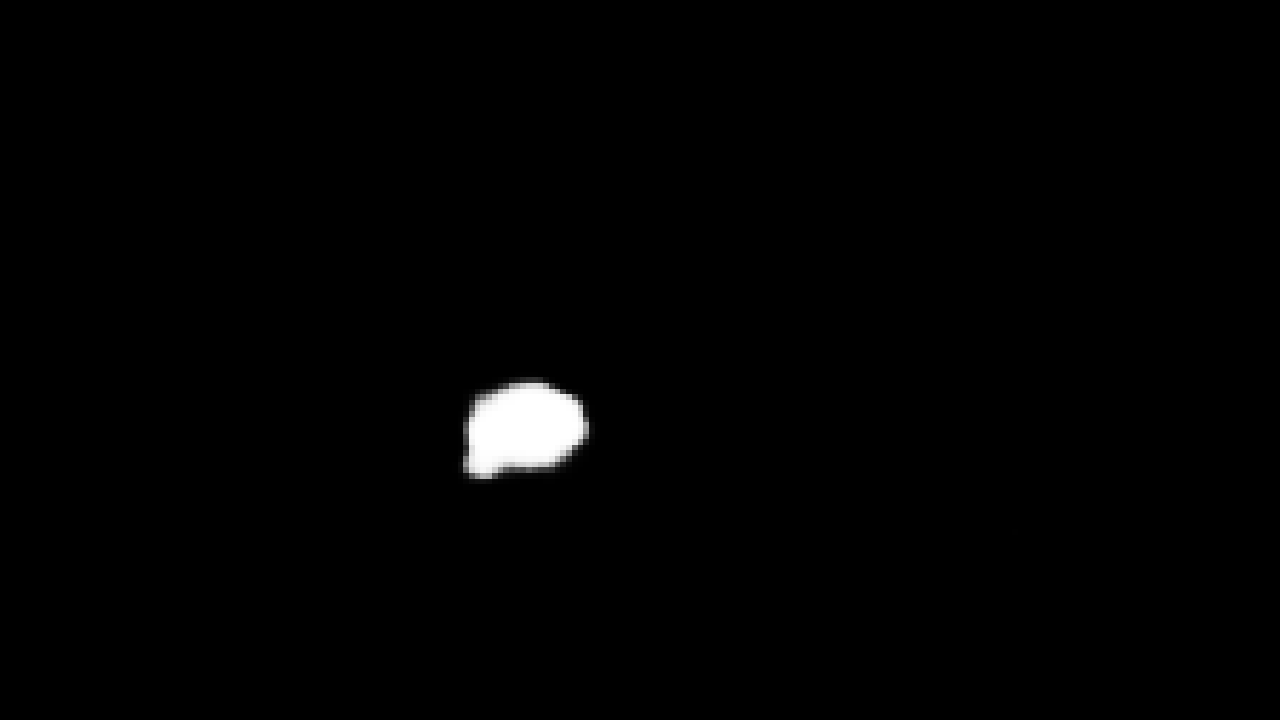}
    \includegraphics[width=0.120\textwidth]{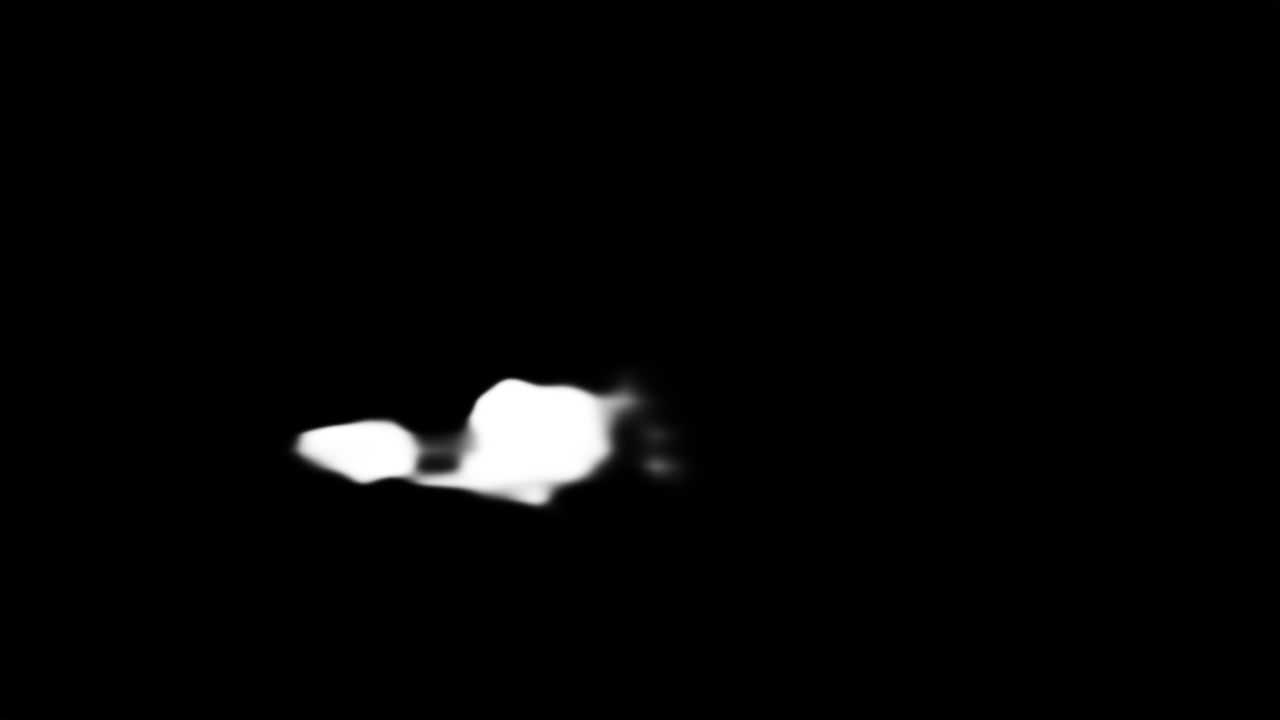}
    \includegraphics[width=0.120\textwidth]{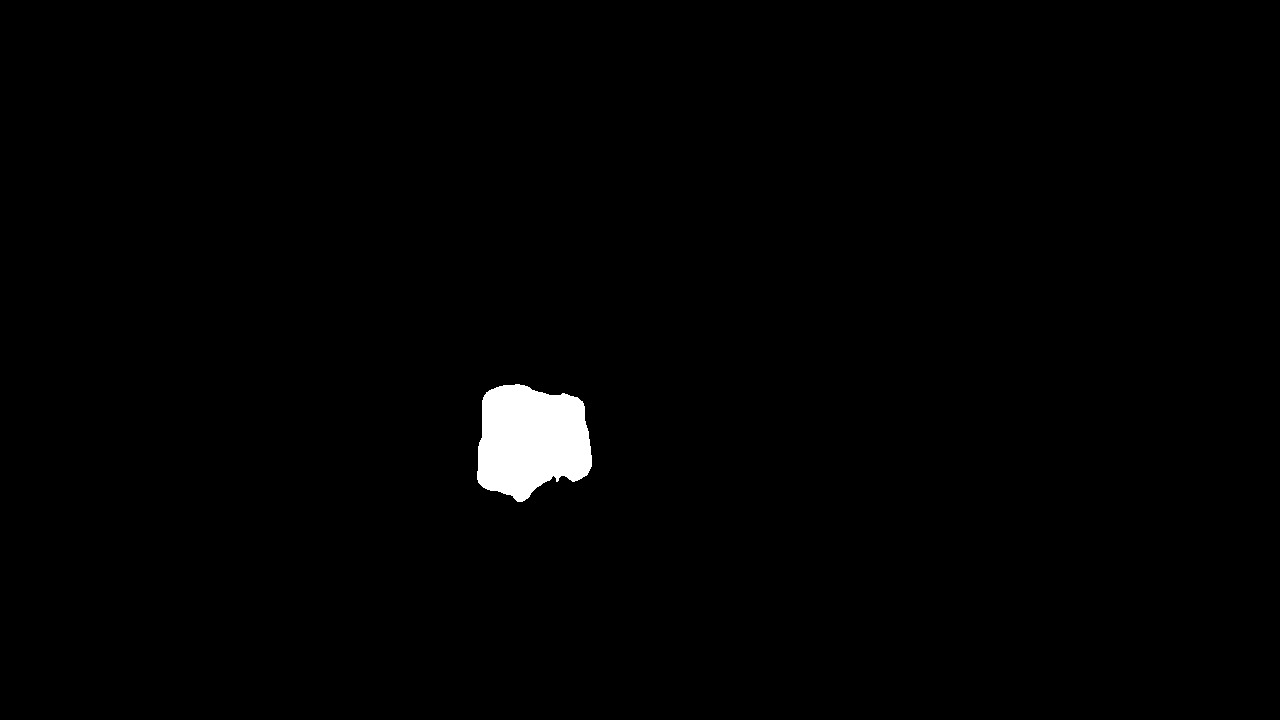}
    \includegraphics[width=0.120\textwidth]{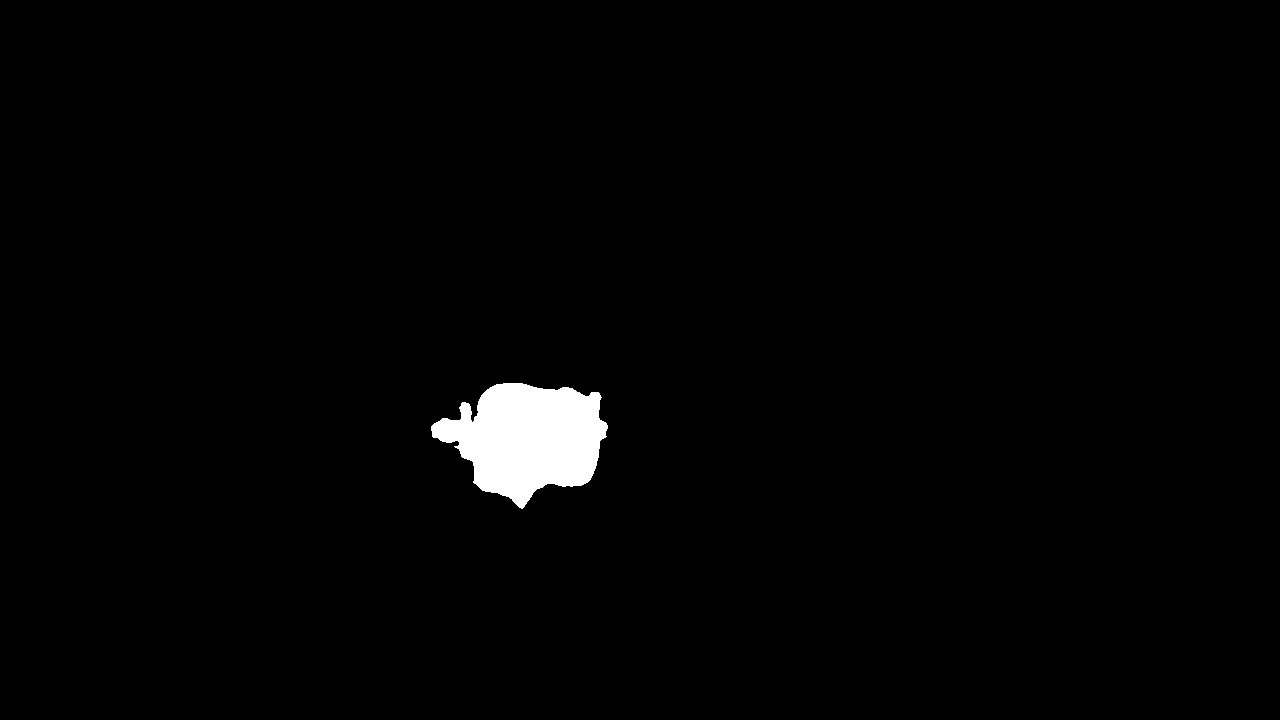}
    \\
    \includegraphics[width=0.120\textwidth]{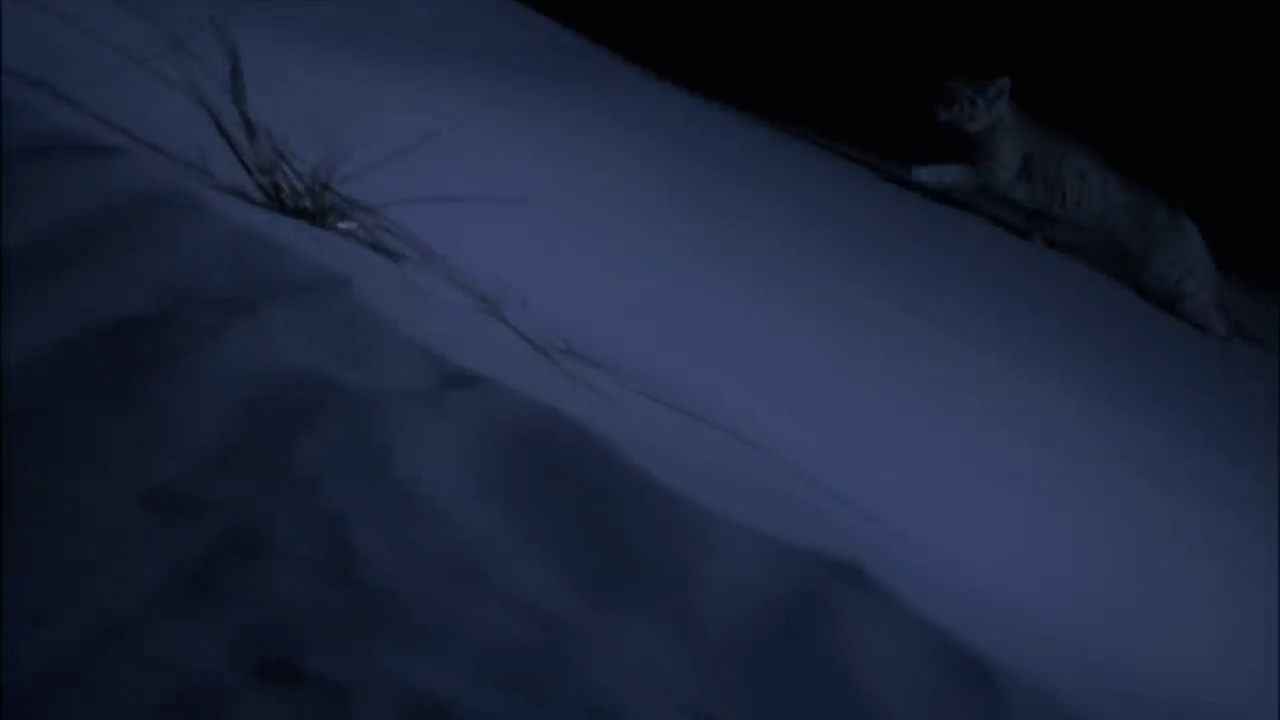}
    \includegraphics[width=0.120\textwidth]{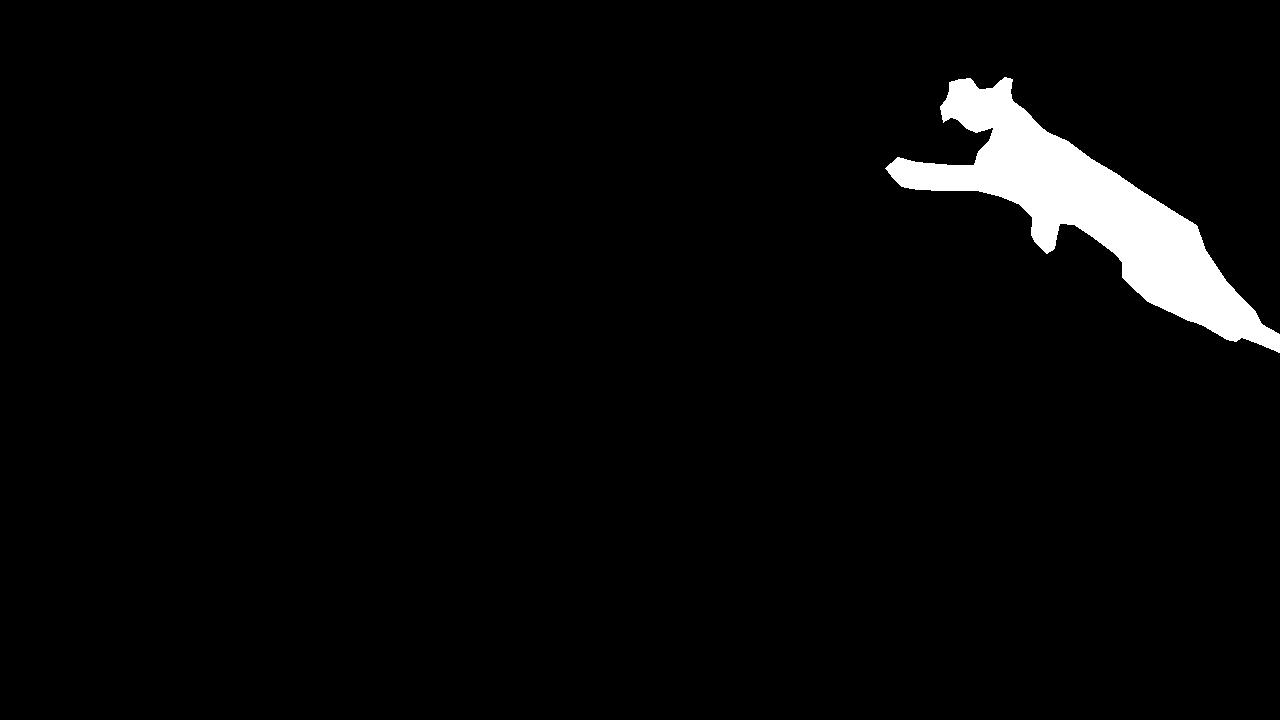}
    \includegraphics[width=0.120\textwidth]{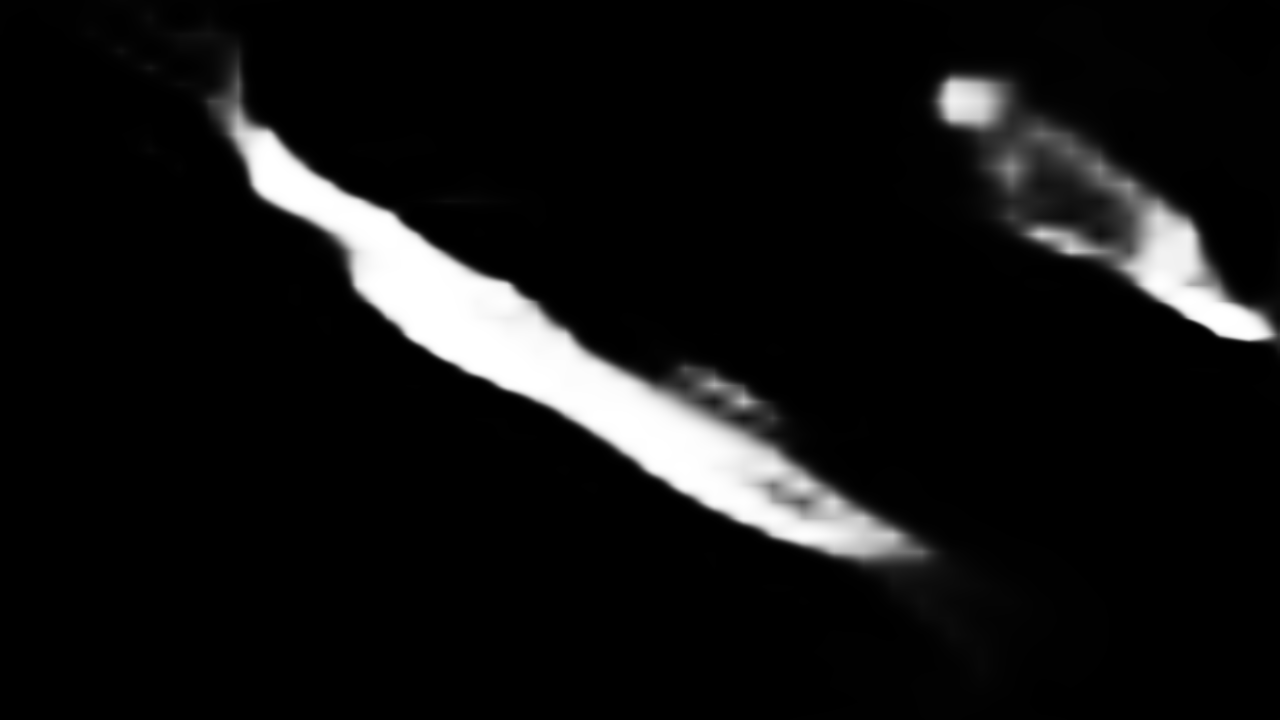}
    \includegraphics[width=0.120\textwidth]{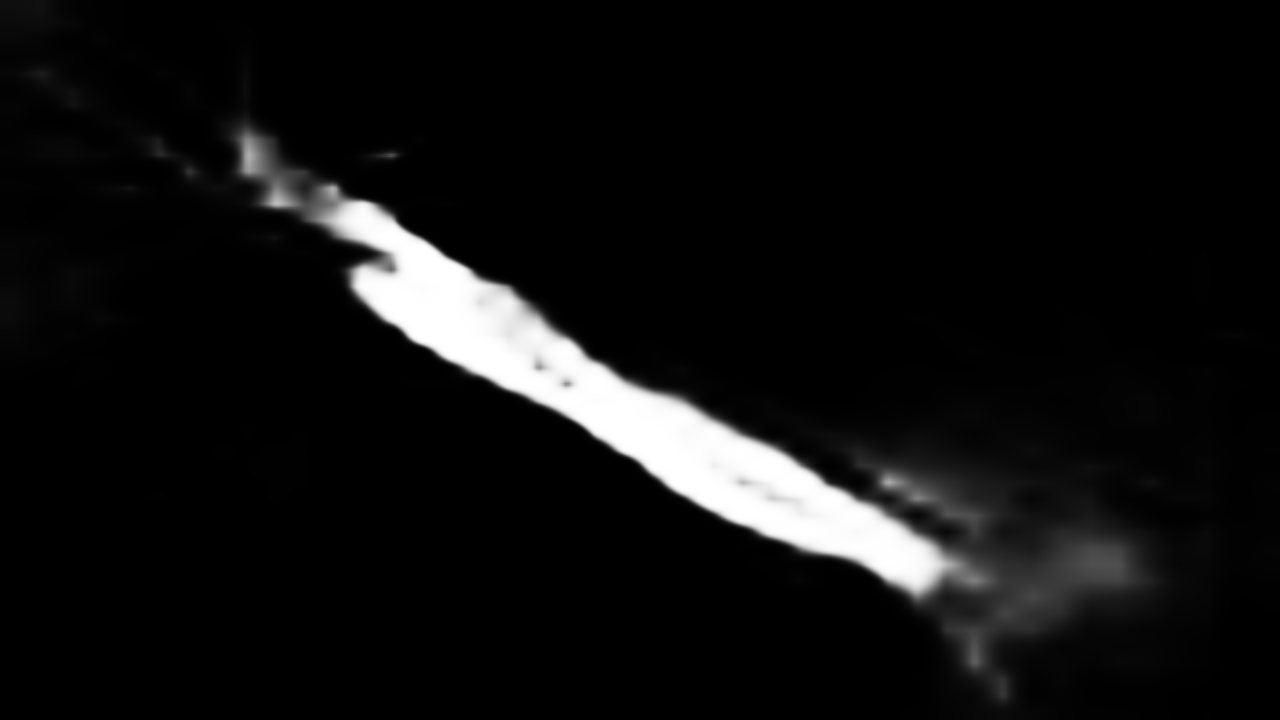}
    \includegraphics[width=0.120\textwidth]{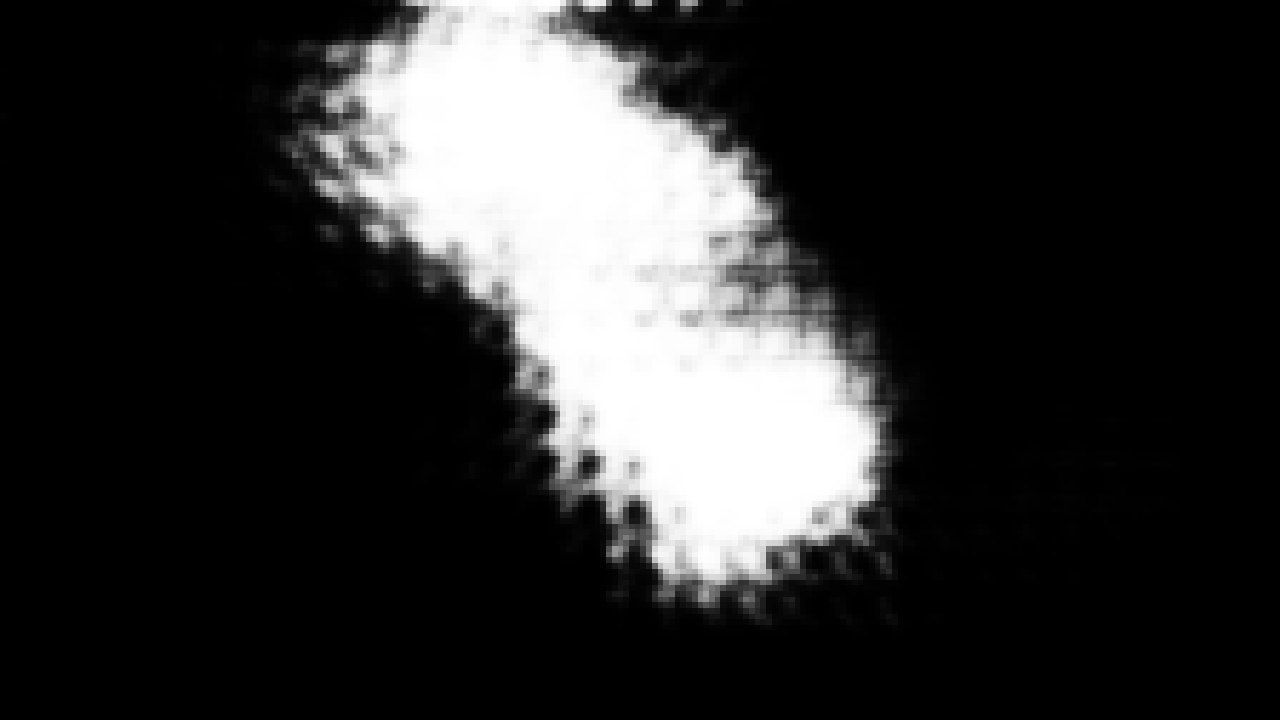}
    \includegraphics[width=0.120\textwidth]{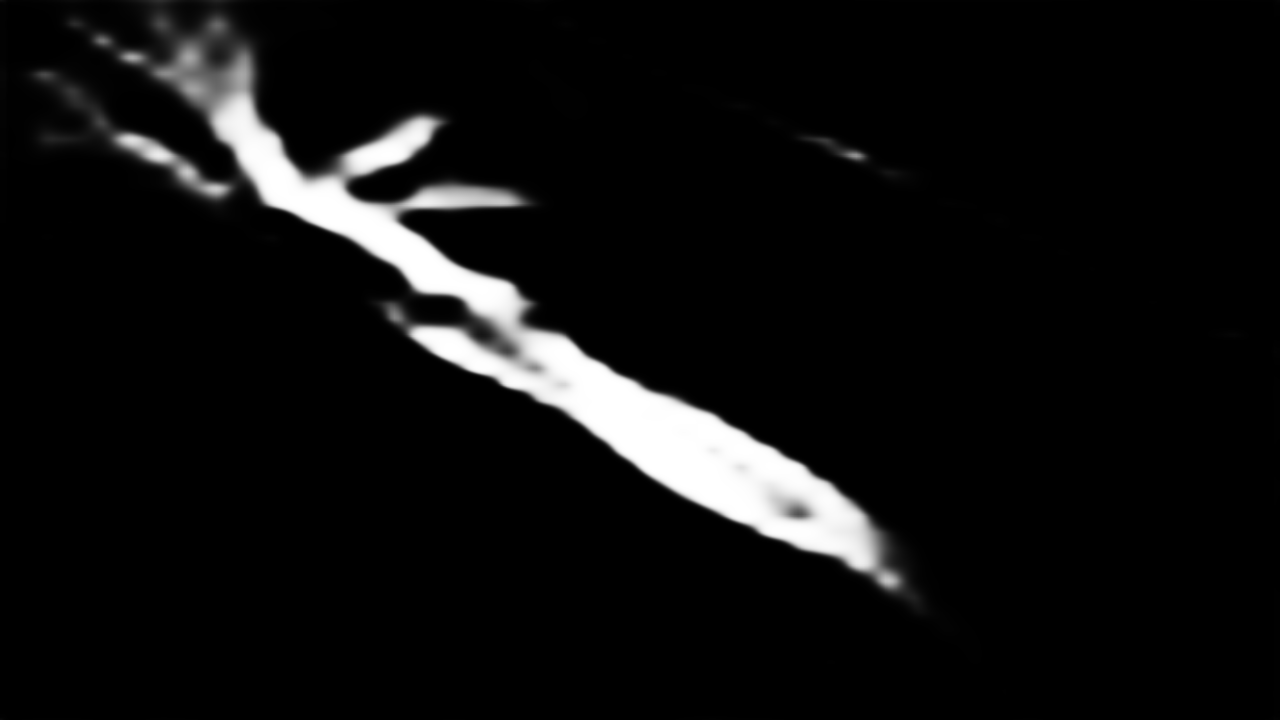}
    \includegraphics[width=0.120\textwidth]{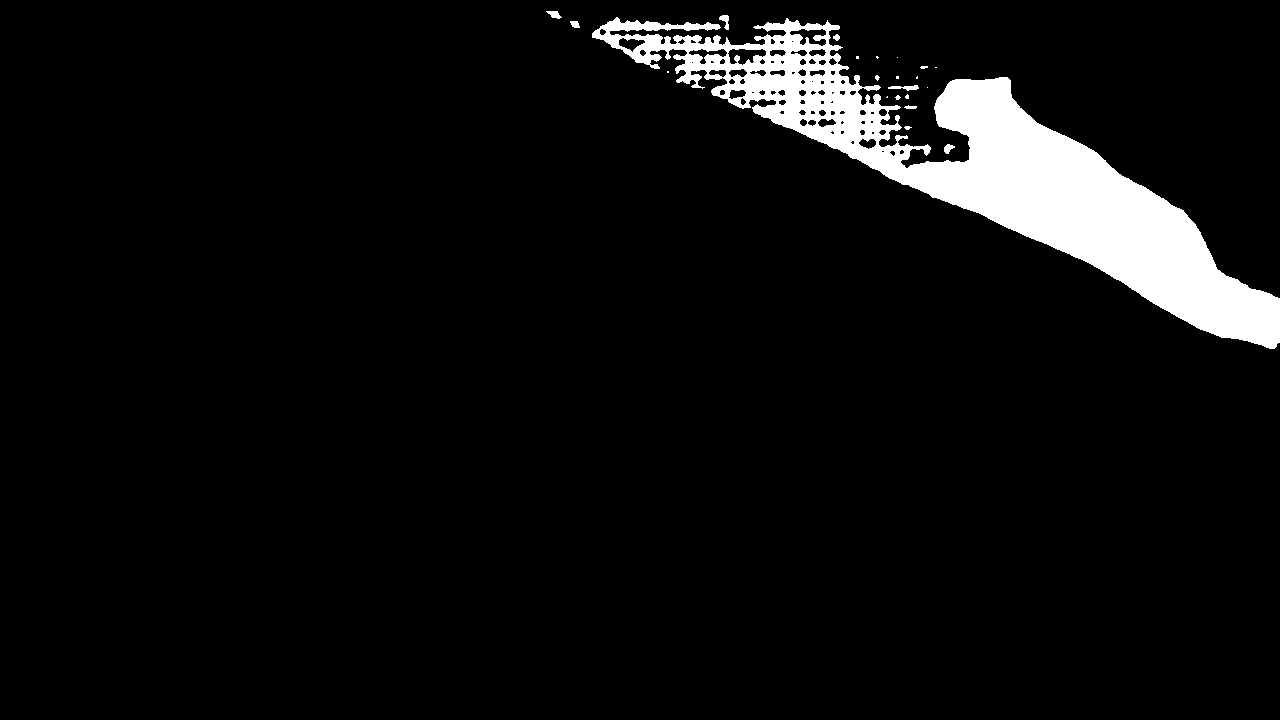}
    \includegraphics[width=0.120\textwidth]{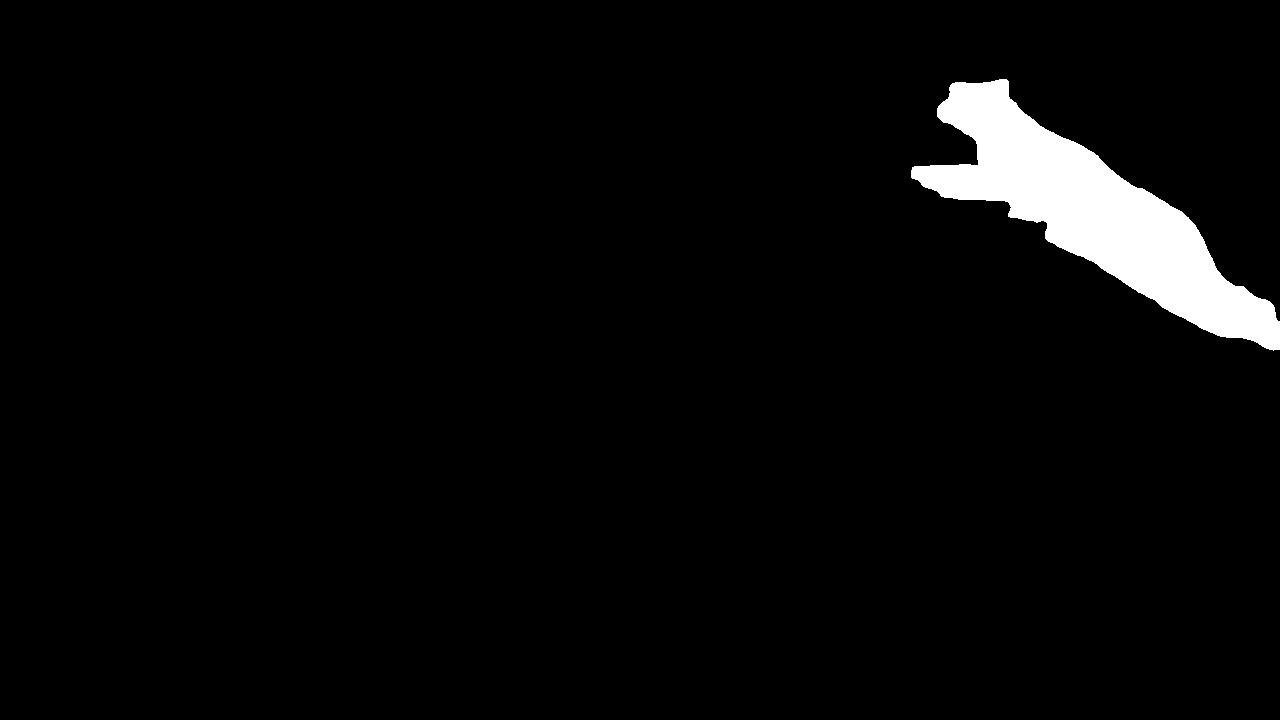}
    \\
    \makebox[0.120\textwidth]{\scriptsize (a) Input Image}
    \makebox[0.120\textwidth]{\scriptsize (b) GT}
    \makebox[0.120\textwidth]{\scriptsize (c) SINet\cite{fan2020camouflaged}}
    \makebox[0.120\textwidth]{\scriptsize (d) SINet\textunderscore v2 \cite{fan2022concealed}}
    \makebox[0.120\textwidth]{\scriptsize (a) MG\cite{yang2021self}}
    \makebox[0.120\textwidth]{\scriptsize (b) SLT-Net\cite{cheng2022implicit}}
    \makebox[0.120\textwidth]{\scriptsize (c) SAM\cite{kirillov2023segment}}
    \makebox[0.120\textwidth]{\scriptsize (d) Ours}
    \\
    \caption{Visual comparison of some recent VCOD models with ours on MoCA-Mask} 
    \label{fig:figure4}
\end{figure*}
 
\begin{figure*}[tb] \centering
    \includegraphics[width=0.120\textwidth]{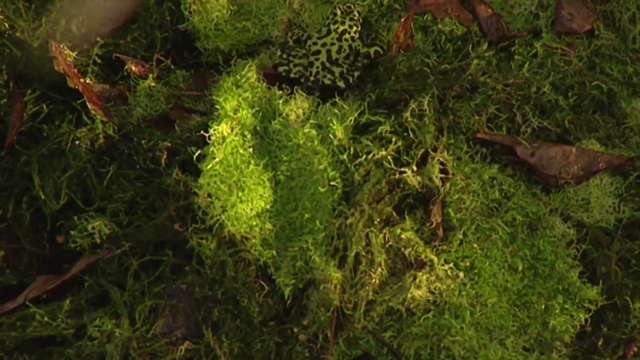}
    \includegraphics[width=0.120\textwidth]{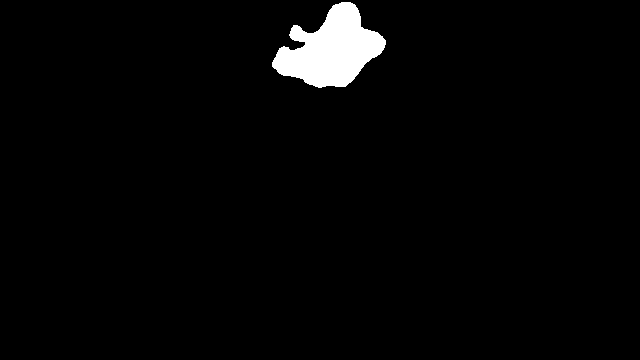}
    \includegraphics[width=0.120\textwidth]{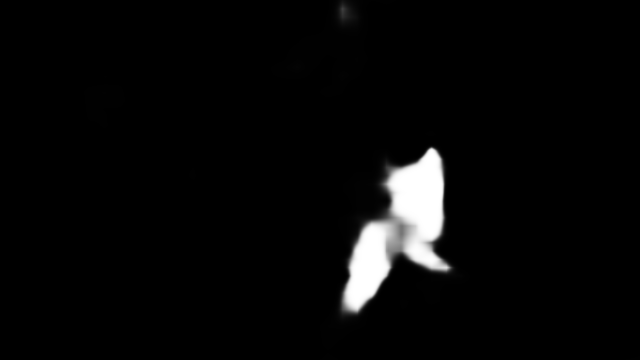}
    \includegraphics[width=0.120\textwidth]{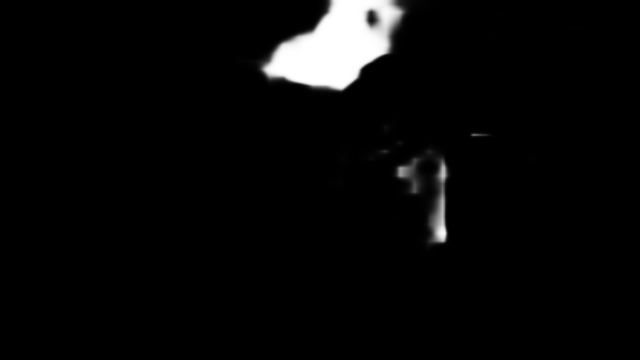}
    \includegraphics[width=0.120\textwidth]{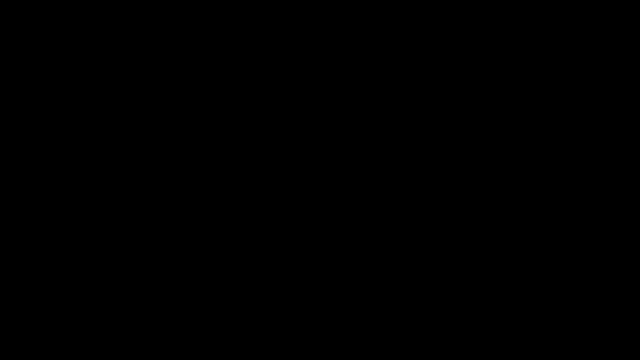}
    \includegraphics[width=0.120\textwidth]{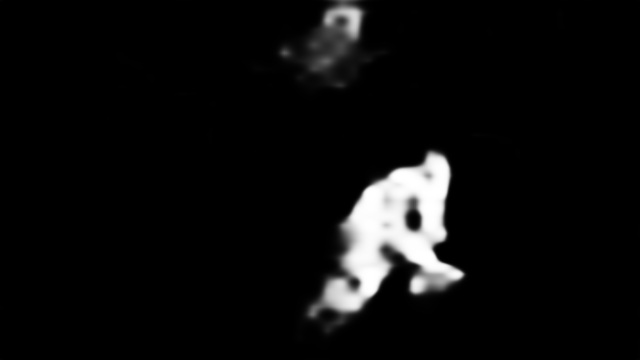}
    \includegraphics[width=0.120\textwidth]{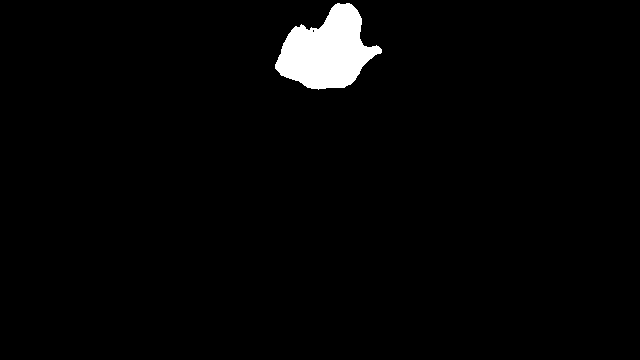}
    \includegraphics[width=0.120\textwidth]{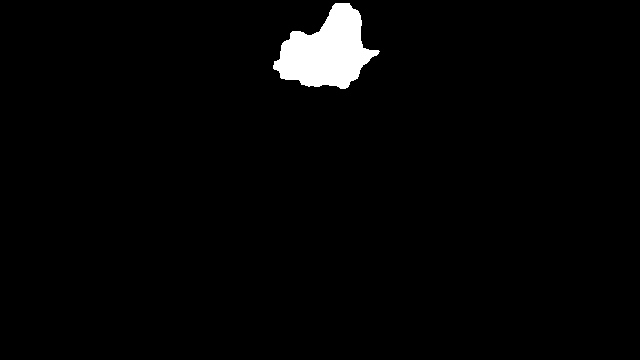}
    \\

    \includegraphics[width=0.120\textwidth]{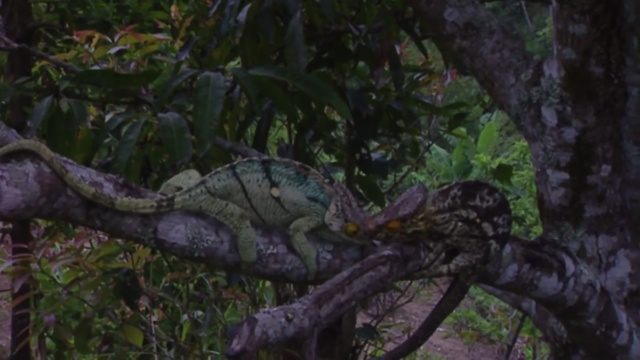}
    \includegraphics[width=0.120\textwidth]{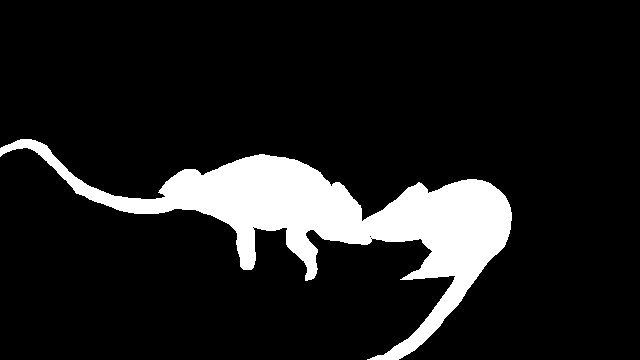}
    \includegraphics[width=0.120\textwidth]{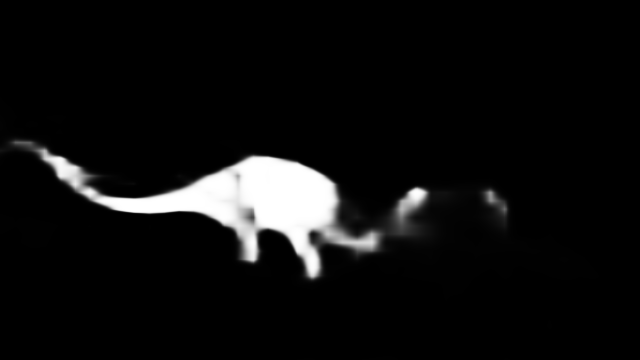}
    \includegraphics[width=0.120\textwidth]{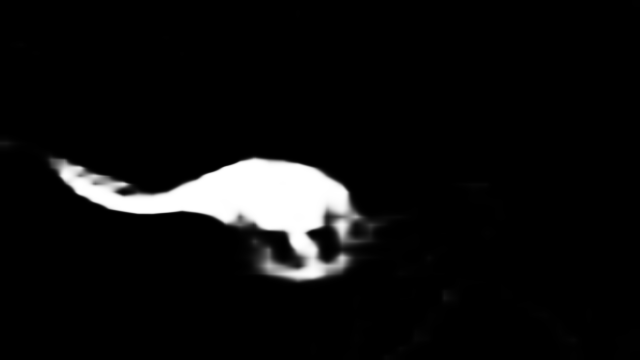}
    \includegraphics[width=0.120\textwidth]{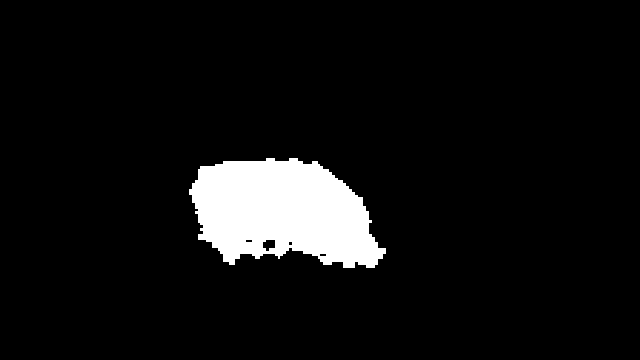}
    \includegraphics[width=0.120\textwidth]{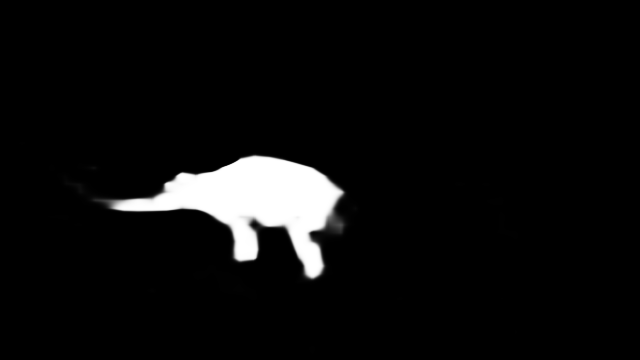}
    \includegraphics[width=0.120\textwidth]{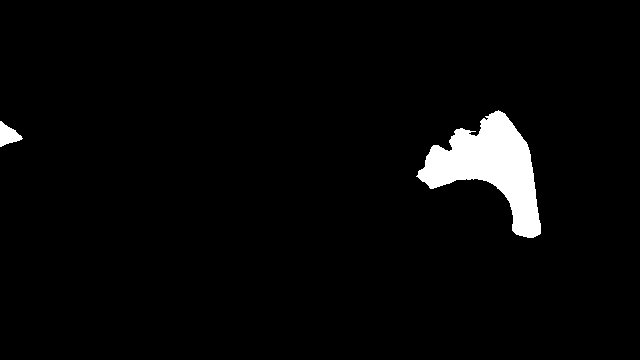}
    \includegraphics[width=0.120\textwidth]{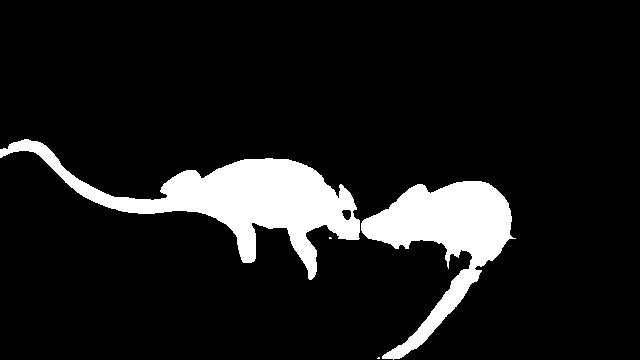}
    \\
    \makebox[0.120\textwidth]{\scriptsize (a) Input Image}
    \makebox[0.120\textwidth]{\scriptsize (b) GT}
    \makebox[0.120\textwidth]{\scriptsize (c) SINet\cite{fan2020camouflaged}}
    \makebox[0.120\textwidth]{\scriptsize (d) SINet\textunderscore v2\cite{fan2022concealed}}
    \makebox[0.120\textwidth]{\scriptsize (a) MG\cite{yang2021self}}
    \makebox[0.120\textwidth]{\scriptsize (b) SLT-Net\cite{cheng2022implicit}}
    \makebox[0.120\textwidth]{\scriptsize (c) SAM\cite{kirillov2023segment}}
    \makebox[0.120\textwidth]{\scriptsize (d) Ours}
    \\
    \caption{Visual comparison of some recent VCOD models with ours on CAD Dataset.} 
    \label{fig:figure5}
\end{figure*}
\begin{multicols}{2}

\subsection{Evaluation Metrics} Six commonly used metrics are employed for VCOD tasks, which include mean absolute error \textit{M} which evaluates the pixel level accuracy between prediction and labeled masks,  Weighted F-measure \textit{${F_{\beta}^{w}}$} \cite{margolin2014evaluate}, Enhanced-alignment measure \textit{$E_{\phi}$} \cite{fan2018enhanced} which simultaneously assesses pixel-level correspondence and overall image statistical alignment, structure measure \textit{$S_{\alpha}$} \cite{fan2017structure} which evaluates region-aware and object-aware structural similarity, mean Dice and mean IoU measures the similarity and overlapping between two sets of data and masks respectively. Larger \textit{${F_{\beta}^{w}}$}, \textit{$E_{\phi}$}, \textit{$S_{\alpha}$}, mDice, mIoU and smaller \textit{ M }indicate better segmentation performance. 

\subsection{Implementation Details}
The images are resized to 1024 x 1024 for static image pre-training and main training after applying RandomAffine, Gaussian Blur, ColorJitter, Random Horizontal Flip, and Random GrayScale augmentation. We use SAM-L with a frozen architecture, where only the PM is trainable. We use a combination of Focal loss \cite{lin2017focal}, Dice Loss \cite{milletari2016v}, and MSE Loss between the IoU prediction of SAM and the IoU of the predicted mask with the ground truth mask in a ratio of 20:1:1 following \cite{kirillov2023segment}.

We adopt a two-stage training approach to ensure stability throughout the training process. First, we pre-train the model on the COD10K dataset, where we randomly pick three frames following \cite{liu2022learning} to form a training sample for the main training of the propagation module. All the training was done with A40 GPU. For optimization, we use AdamW \cite{kingma2014adam, loshchilov2017decoupled} with a weight decay of 1e-2 and a batch size of 4 for both stages. The pre-training is done for 7.6k iterations with an initial learning rate of 4e-4, and the main training is performed for 2.5k iterations with an initial learning rate of 5e-5. We employ a multi-step learning rate decay schedule, reducing the learning rate by half after 3.8k iterations during the pre-training stage.

We initiate the process during inference with the first frame and its corresponding ground truth mask. Subsequent masks are predicted by concatenating all frames and masks in the memory with a set memory length of 2. To manage memory, we remove the oldest frame whenever the memory reaches full capacity.

\subsection{Results}
Table \ref{tab:quant_results} illustrates our model performance with recent state-of-the-art VCOD methods. For the evaluation of SAM, we provide the geometric center of the animal as a point prompt, which is sampled from the ground truth. Since SAM-PM utilizes the ground truth of the initial frame as input to the model, we exclude it from the metric calculations.

We provide qualitative results on the MoCA-Mask and CAD datasets in Figure \ref{fig:figure4} and Figure \ref{fig:figure5}, respectively. When compared with the previous SOTA SLT-Net, our method achieves average performance gains of  82.31\%, 15.37\%, 7.11\%, 200\%, 65\%, and 84.55\% in terms of \textit{${F_{\beta}^{w}}$}, \textit{$S_{\alpha}$}, \textit{$E_{\phi}$}, \textit{M}, mDic, mIoU respectively on MoCA-Mask dataset.

This demonstrates that our model can segment out accurately and locate camouflaged objects in various complex and challenging scenarios. Our method outperforms all the previous methods on the CAD dataset. It surpasses the previous SOTA SLT-Net by 25.16\%, 4.74\%,  66.67\%, 20.48\% and 22.94\% in terms of \textit{${F_{\beta}^{w}}$}, \textit{$S_{\alpha}$}, \textit{M}, mDic, mIoU respectively.

\subsection{Ablation}
We perform ablation on the MoCA-Mask and CAD dataset to analyze the performance contribution of positional embedding and affinity in the propagation framework. The results are tabulated in table \ref{tab:abellation}.

\textbf{Positional Embedding}. Incorporating positional embedding contributes to a performance boost of 5.5\% on \textit{$S_{\alpha}$}, 19.11\% on \textit{${F_{\beta}^{w}}$} , 4.77\% on \textit{$E_{\phi}$}, 33.33\% on \textit{M}, 18.32\% on mDic, and 19.5\% on mIoU for MoCA-Mask Dataset and a performance boost of 2.39\% on \textit{$S_{\alpha}$}, 5.24\% on \textit{${F_{\beta}^{w}}$} , 1.08\% on \textit{$E_{\phi}$}, 22.22\% on \textit{M}, 6.07\% on mDic and 7.87\% on mIoU for CAD dataset. 
The positional embedding is introduced to preserve spatial information in embedding generated by the SAM encoder. Camouflaged images often involve intricate patterns and subtle variations in colour and texture. Preserving the spatial arrangement of these patterns is crucial for distinguishing between the camouflage and the background. The patterns are designed to disrupt the perception of shapes and structures. If a model treats different spatial arrangements as equivalent (symmetric), it may struggle to differentiate between the camouflage and the background, which is now solved by using positional embedding.
\\

\textbf{MPAM}. Training the model with MPAM results in a performance increase of 10.13\% on \textit{$S_{\alpha}$}, 33.72\% on \textit{${F_{\beta}^{w}}$} , 20.8\% on \textit{$E_{\phi}$}, 22.22\% on \textit{M} , 35\% on mDic and 36.78\% on mIoU for MoCA-Mask dataset and a performance increase of 8.32\% on \textit{$S_{\alpha}$}, 22.61\% on \textit{${F_{\beta}^{w}}$} , 18.97\% on \textit{$E_{\phi}$}, 27.78\% on \textit{M} , 25.85\% on mDic and 29.05\% on mIoU for CAD dataset. It is a crucial module in refining the superficial predictions from the previous module. Camouflaged images tend to produce low-confidence fuzzy features, which are now rectified with the help of the affinity model, leading to improvement in performance.

\section{Limitations}
Even though the PM is extremely small and efficient in terms of parameters, the throughput and latency of SAM-PM are heavily bottlenecked due to the large encoder (contributing to more than \>98\% parameters in SAM-PM) that is being used. We have also observed the model failing sometimes for smaller targets. Since the encoder is one of the crucial components in effectively getting the essential target representation, our PM is limited by the encoder's performance in COD tasks. Our SAM-PM approach is directed towards parameter-efficient fine-tuning therefore, further encoder fine-tuning might be essential for better performance. Currently, we employ a naive queue-based memory bank, which can be improved further by using complex strategies, such as memory updation and contrastive items in CTVIS \cite{ying2023ctvis}.

\end{multicols}
\begin{table*}[tb]
    \caption{Ablation results on different VCOD Datasets.}
    \label{tab:abellation}
    \centering
    \resizebox{1.0\textwidth}{!}{
    \begin{tabular}{ |l|*{6}{c}|*{6}{c}| }
    \hline
    & \multicolumn{6}{c|}{MoCA-Mask w/o pseudo labels} &  \multicolumn{6}{c|}{CAD}  \\
    \cline{2-13}
    Model & \textit{$S_{\alpha}\uparrow$} & \textit{${F_{\beta}^{w}}\uparrow$} & \textit{$E_{\phi}\uparrow$} & \textit{$M\downarrow$} & mDic & mIoU & \textit{$S_{\alpha}\uparrow$} & \textit{${F_{\beta}^{w}}\uparrow$} & \textit{$E_{\phi}\uparrow$} & \textit{$M\downarrow$} & mDic & mIoU  \\
    \hline
    
    SAM-PM & \textbf{0.728} & \textbf{0.567} & \textbf{0.813} & \textbf{0.009} & \textbf{0.594} & \textbf{0.502} & \textbf{0.729} & \textbf{0.602} & \textbf{0.746} & \textbf{0.018} & \textbf{0.594} & \textbf{0.493} \\
    SAM-PM without positional embeddings & 0.690 & 0.476 & 0.776 & 0.012 & 0.502 & 0.420 & 0.712 & 0.572 & 0.738 & 0.022 & 0.560 & 0.457 \\
    SAM-PM without MPAM & 0.661 & 0.424 & 0.673 & 0.011 & 0.440 & 0.367 & 0.673 & 0.491 & 0.627 & 0.023 & 0.472 & 0.382 \\
    \hline
    \end{tabular} }
    
\end{table*}

\begin{multicols}{2}

\section{Conclusion}
In summary, our novel SAM Propagation Module (SAM-PM) presents a noteworthy improvement in Video Camouflaged Object Detection, outperforming previous state-of-the-art models. Our Propagation Module seamlessly integrates with the Segment Anything Model (SAM), requiring less than 1\% parameters for training. Its successful integration proves its effectiveness in tackling camouflaged objects in videos and suggests its potential to enhance large foundation models for real-world applications. The demonstrated performance improvements and minimal overhead of SAM-PM offer a promising avenue for advancing the field, opening possibilities for more efficient and accurate video-based segmentation tasks.